\documentclass[journal]{IEEEtran}
\usepackage{cite}
\usepackage{graphicx}
\usepackage{amsmath}
\usepackage{array}
\usepackage{amssymb}
\usepackage{booktabs}
\usepackage{bbding}
\usepackage{subfigure}
\usepackage{textcomp}
\usepackage{todonotes}
\usepackage{gensymb}
\usepackage{xcolor}
\usepackage{color}
\usepackage{url}
\usepackage{multirow}
\usepackage{mathrsfs}
\usepackage{booktabs, makecell}
\usepackage{threeparttable}
\usepackage{epsfig}
\usepackage{algorithmicx,algorithm}
\usepackage{algpseudocode}
\usepackage{bbm}

\usepackage{titlesec}

\def\ie{{\em i.e.}}
\def\eg{{\em e.g.}}
\def\etal{{\em et al.}}

\usepackage{diagbox}
\begin{document}
\title{Surveillance Face Anti-spoofing}
\author{
	Hao Fang, 
    Ajian Liu, 
	Jun Wan,~\IEEEmembership{Senior Member,~IEEE},
	Sergio Escalera,~\IEEEmembership{Senior Member,~IEEE} \\
	Chenxu Zhao, 
	Xu Zhang, 
	Stan Z. Li,~\IEEEmembership{Fellow ,~IEEE}, 
    Zhen Lei,~\IEEEmembership{Senior Member,~IEEE}
	\thanks{
 Corresponding author: Jun Wan (e-mail: jun.wan@ia.ac.cn).\\
 Hao Fang, Ajian Liu, Jun Wan and Zhen Lei are with the National Laboratory of Pattern Recognition (NLPR), Institute of Automation Chinese Academy of Sciences (CASIA) and School of Artificial Intelligence, University of Chinese Academy of Sciences (UCAS), Beijing, China (e-mail: \{fanghao2021, ajian.liu, jun.wan, zhen.lei\}@ia.ac.cn).  \\
Xu Zhang is with Beijing Normal University School of Artificial Intelligence, Beijing, China (e-mail: xuzhang0908@mail.bnu.edu.cn). \\
Chenxu Zhao is with the SailYond Technology, Beijing, China ( e-mail: zhaochenxu@sailyond.com). \\
Sergio Escalera is with the Universitat de Barcelona (UB), Barcelona, Computer Vision Center (CVC), and Aalborg University (AAU) (e-mail: sergio@maia.ub.es). \\
Stan Z. Li is with Westlake University, Hangzhou, China (e-mail: stan.zq.li@westlake.edu.cn).	
 }
}
% \markboth{Journal of IEEE Transactions on Information Forensics and Security,~Vol.~xx, No.~x, xxx~2022}
% {Fang \MakeLowercase{~\etal}: Surveillance Face Anti-spoofing Recognition}
\maketitle

\begin{abstract}
Face Anti-spoofing (FAS) is essential to secure face recognition systems from various physical attacks. However, recent research generally focuses on short-distance applications (\ie, phone unlocking) while lacking consideration of long-distance scenes (\ie, surveillance security checks). In order to promote relevant research and fill this gap in the community, 
we collect a large-scale \emph{Su}rveillance \emph{Hi}gh-\emph{Fi}delity \emph{Mask} (SuHiFiMask) dataset captured under $40$ surveillance scenes, which has $101$ subjects from different age groups with $232$ 3D attacks (high-fidelity masks), $200$ 2D attacks (posters, portraits, and screens), and $2$ adversarial attacks. In this scene, low image resolution and noise interference are new challenges faced in surveillance FAS. Together with the SuHiFiMask dataset, we propose a Contrastive Quality-Invariance Learning (CQIL) network to alleviate the performance degradation caused by image quality from three aspects: (1) An Image Quality Variable module (IQV) is introduced to recover image information associated with discrimination by combining the super-resolution network. (2) Using generated sample pairs to simulate quality variance distributions to help contrastive learning strategies obtain robust feature representation under quality variation. (3) A Separate Quality Network (SQN) is designed to learn discriminative features independent of image quality. Finally, a large number of experiments verify the quality of the SuHiFiMask dataset and the superiority of the proposed CQIL.

\end{abstract}
\IEEEpeerreviewmaketitle

% BODY TEXT
\section{Introduction}

\IEEEPARstart{F}{ace} Presentation Attack Detection (PAD) technology is a crucial step to enhance the security of face recognition systems and plays an increasingly important role in resisting malicious attacks, such as print-attack~\cite{zhang2012face}, replay-attack~\cite{chingovska2012effectiveness}, or face-mask~\cite{nesli2013spoofing}. Although current works~\cite{Liu2018Learning, shao2019multi,george2019deep,yu2020searching,zhang2020face,liu2020disentangling,yang2021few,chen2021generalizable,liu2021face,li20203dpc} have achieved satisfactory performance in short-distance applications, such as phone unlocking, face payment, and access authentication, they are still sensitive to face quality and fail in long-distance applications, which hinders the expansion of FAS to surveillance scenarios.
\begin{figure}[t]
	\centering
    	\includegraphics[width=1.0\linewidth,height=0.35\textwidth]{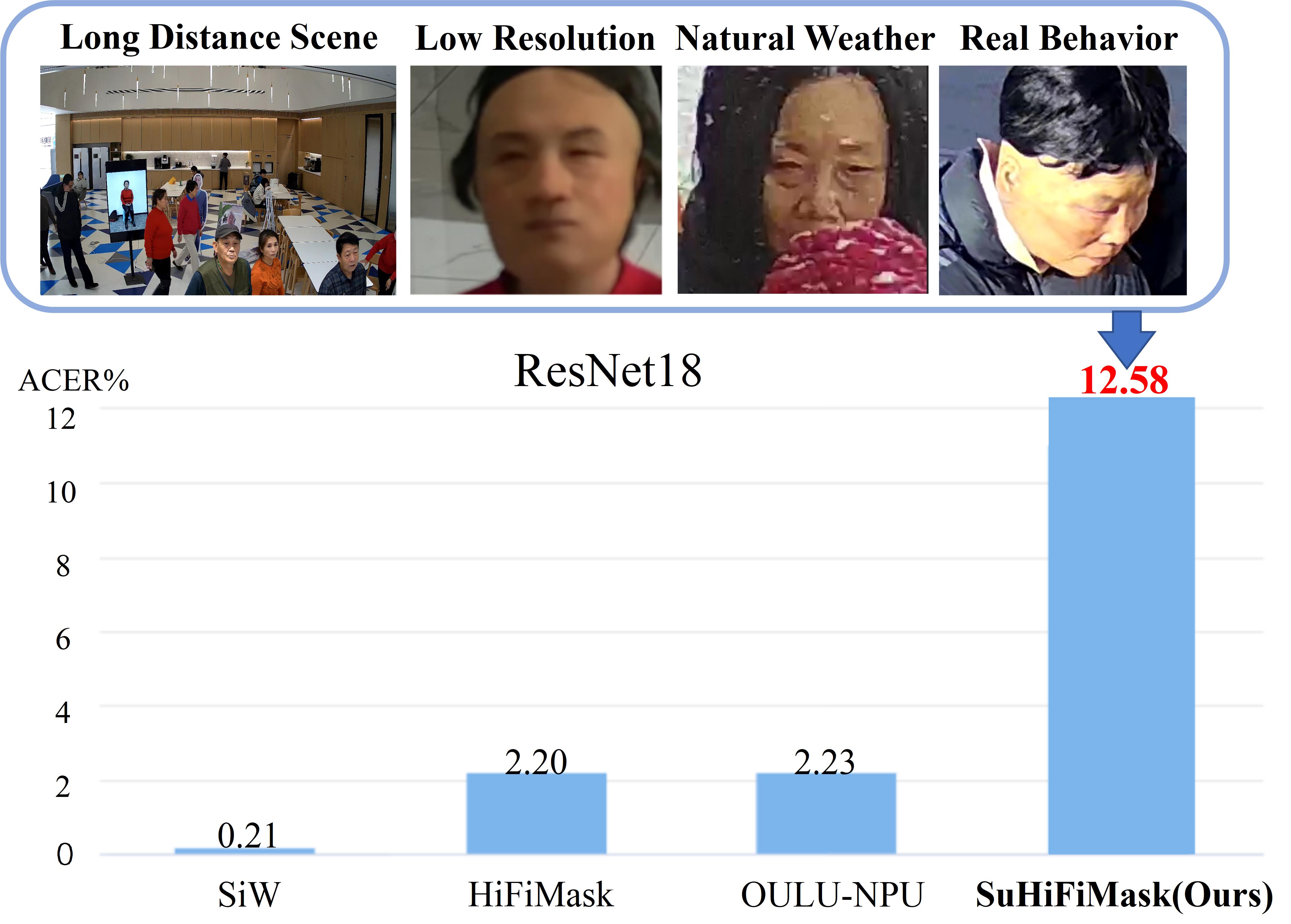}
	\vspace{-0.4cm}
	\caption{Performance comparisons on the SiW, HiFiMask, OULU-NPU, and proposed SuHiFiMask dataset using the same ResNet18 network. It shows significant performance degradation under surveillance FAS. }
	\label{fig:feature}
\end{figure}

With the popularity of remote cameras and the improvement of surveillance networks, the development of smart cities has put forward higher requirements for traditional visual technologies in surveillance. Benefited from the release of face recognition datasets~\cite{cheng2018surveillance,nada2018pushing,grgic2011scface} in the surveillance scene and driven by related algorithms~\cite{Kim_2022_CVPR,8600370,zhong2021sface}, the face recognition system has gradually got rid of the constraint of verification distance, and can use the surveillance camera to complete real-time capture, self-service access control, and self-service supermarket payment. However, the FAS community is still stuck in the protection of the face recognition system under short-distance conditions, and cannot serve for the detection of spoofing faces under a long-distance natural behavior. We analyze two reasons that hinder the development of PAD technologies: (1) \textbf{Lack of a dataset that can truly simulate the attack in surveillance.} The existing FAS datasets, whether 2D print or replay attacks~\cite{Boulkenafet2017OULU,Liu2018Learning,zhang2020casia}, or 3D mask attacks~\cite{nesli2013spoofing,galbally2016three,steiner2016reliable,liu20163d2,liu2022contrastive}, require the subjects to face the acquisition device under distance constraints. However, diversified surveillance scenes, rich spoofing types, and natural human behavior are important assessment factors for the surveillance FAS dataset collection. (2) \textbf{Low-quality faces in the surveillance scenarios cannot meet the requirements of fine-grained feature-based FAS tasks.} The existing FAS algorithms, whether based on color-texture feature learning~\cite{yu2020fas,jia2020survey,jia20203d,george2019biometric}, face depth structure fitting~\cite{Liu2018Learning,george2019deep}, or remote photoplethysmography (rPPG)-based detection~\cite{liu2018remote,lin2019face,liu2021multi}, require high-quality image details to ensure high performances. As illustrated in Fig.~\ref{fig:feature}, the resolution of faces under long-distance surveillance is small and contains noise from motion blur, occlusion, bad weather, and other bad factors. These are new challenges for algorithm design in surveillance FAS. 
%QMUL-SURVFACE、Scface、IJB-C

\begin{table*}[ht]
\caption{Comparison of public face anti-spoofing datasets. * notes that SuHiFiMask is focused on surveillance scenes where both real people and fake attacks appear at the same time. Therefore, the number of real videos and the number of fake videos are both 10195. }
\begin{threeparttable}
\scalebox{0.74}{
\begin{tabular}{|c|c|c|c|c|c|c|c|c|}
\hline
Dataset, Year                                                      & \#sub.  & \begin{tabular}[c]{@{}c@{}}Distance\\ (Long/Short)\end{tabular} & Materials  & Scenes                 & Light, Weather                      & Attacks     & Devices                                                                                                                                           & \begin{tabular}[c]{@{}c@{}}\#Videos\\ (\#Live/\#Fake)\end{tabular} \\ \hline
3DMAD~\cite{erdogmus2013spoofing}, 2013                                                        & 17  & Short   & Paper, Resin                                                        & Constrained scenes                                                                                                    & Adjustment                                                                                                       & 2D/2.5D image                                                                                         & Kinect                                                                                                                                            & 255(170/85)                                                      \\ \hline
3DFS-DB~\cite{galbally2016three}, 2016                                                      & 26  & Short   & Plastic                                                                 & Office                                                                                                         & Adjustment                                                                                                       & 2D/2.5D image, 3D Mask                                                                                            & Kinect, Carmine 1.09                                                                                                                               & 520(260/260)                                                     \\ \hline
BRSU~\cite{steiner2016reliable}, 2016                                                         & 137  & Short  & \begin{tabular}[c]{@{}c@{}}Silicone, Plastic,\\ Resin, Latex\end{tabular} & \begin{tabular}[c]{@{}c@{}}Disguise,\\ Counterfeiting\end{tabular}                                              & Adjustment                                                                                                       & \begin{tabular}[c]{@{}c@{}} 2D image\end{tabular}                                             & SWIR, Color                                                                                                                                        & 141(0/141)                                                       \\ \hline
MARsV2~\cite{liu20163d}, 2016                                                       & 12  & Short  & \begin{tabular}[c]{@{}c@{}}ThatsMyFace,\\ REAL-F\end{tabular}            & Office                                                                                                         & \begin{tabular}[c]{@{}c@{}}Six directions of light\end{tabular} & 3D Mask                                                                                                          & \begin{tabular}[c]{@{}c@{}}Logitech C920, Industrial Cam,\\ EOS M3, Nexu 4, IPhone 6,\\ Samsung S7, Sony Tablet S\end{tabular}                          & \begin{tabular}[c]{@{}c@{}}1008\\ (504/504)\end{tabular}         \\ \hline
SMAD~\cite{manjani2017detecting}, 2017                                                         & Online & Short & Silicone                                                                & -                                                                                                              & Varying light                                                                                                 & 2D image, 3D Mask                                                                                                 & Varying Cam                                                                                                                                       & 130(65/65)                                                       \\ \hline
MLFP~\cite{agarwal2017face}, 2017                                                         & 10  & Short   & Latex, Paper                                                             & Indoor, Outdoor                                                                                                 & Daylight                                                                                                         & \begin{tabular}[c]{@{}c@{}}2D image\end{tabular}                                           & \begin{tabular}[c]{@{}c@{}}Visible,\\ Near infrared, Thermal\end{tabular}                                                                            & \begin{tabular}[c]{@{}c@{}}1350\\ (150/1200)\end{tabular}        \\ \hline
ERPA~\cite{bhattacharjee2017you}, 2017                                                         & 5   & Short   & Resin, Silicone                                                          & Indoor                                                                                                         & Room light                                                                                                       & \begin{tabular}[c]{@{}c@{}} 3D Mask\end{tabular}                                         & Xenic Gobi, Thermal Cam                                                                                                                            & 86                                                               \\ \hline
WMCA~\cite{george2019biometric}, 2019                                                         & 72  & Short  & \begin{tabular}[c]{@{}c@{}}Plastic,\\ Silicone, Paper\end{tabular}        & Indoor                                                                                                         & \begin{tabular}[c]{@{}c@{}}Office/LED/Day light\end{tabular}                                       & \begin{tabular}[c]{@{}c@{}} 2D image, 3D Mask\end{tabular}                               & \begin{tabular}[c]{@{}c@{}}Intel RealSense SR 300,\\ Seek Thermal, Compact PRO\end{tabular}                                                         & \begin{tabular}[c]{@{}c@{}}1670\\ (347/1332)\end{tabular}        \\ \hline
\begin{tabular}[c]{@{}c@{}} 3DMask~\cite{yu2020fas}, 2020\end{tabular} & 48 & Short    & Plaster                                                                 & Indoor, Outdoor                                                                                                 & \begin{tabular}[c]{@{}c@{}}Six directions of light\end{tabular}       & 2D image, 3D Mask                                                                                                & Apple, Huawei, Samsung                                                                                                                              & \begin{tabular}[c]{@{}c@{}}1152\\ (288/864)\end{tabular}         \\ \hline
HiFiMask~\cite{liu2022contrastive}, 2021                                                     & 75 & Short   & \begin{tabular}[c]{@{}c@{}}Transparent,\\ Plaster, Resin\end{tabular}      & \begin{tabular}[c]{@{}c@{}}White, Green,\\ Tricolor, Sunshine,\\ Shadow, Motion\end{tabular}                        & \begin{tabular}[c]{@{}c@{}}Six directions of light\end{tabular}  & 2D image, 3D Mask                                                                                                 & \begin{tabular}[c]{@{}c@{}}IPhone 11, IPhone X,\\ MI10, P40, S20, Vivo, HJIM\end{tabular}                                                               & \begin{tabular}[c]{@{}c@{}}54,600\\ (13,650/40,950)\end{tabular}        \\ \hline
\textbf{SuHiFiMask (ours), 2022}                                                  & 101 & \textbf{Long}   & \begin{tabular}[c]{@{}c@{}}Resin, Plaster,\\ Silicone, Paper\end{tabular}        & \begin{tabular}[c]{@{}c@{}}Security check lane, \\Theater, Parking lot \tnote{1}\end{tabular} & \begin{tabular}[c]{@{}c@{}}Day/Night light, \\ Sunny/Windy/ \\Cloudy/Snowy day \end{tabular}                     & \begin{tabular}[c]{@{}c@{}}2D image, Video replay, \\ 3D Mask \end{tabular} & \begin{tabular}[c]{@{}c@{}} Surveillance cameras\tnote{2}\end{tabular} & \begin{tabular}[c]{@{}c@{}}10,195*\\ (10,195/10,195)\end{tabular}   \\ \hline
\end{tabular}}

\begin{tablenotes}\footnotesize
\item[1] 40 real surveillance environments, including indoor as well as outdoor. Please see Fig.2 in \emph{Appendix} for more details.
\item[2] dahua: DH-IPC-HFW4843M, DH-P80A1-SA; HIKVISION: DS-2CD3T87WD-L, DS-2CD3T86FWDV2-I3S; TP-LINK: TL-IPC586FP, TL-IPC586HP;\\  ZHONGDUN: ZD5920-Gi4N (Brand name: Camera model) . 
\end{tablenotes}
\end{threeparttable}
\label{table:datasets_compare}
\end{table*}

In order to fill the gap in surveillance scenes of the FAS community, we target to solve two challenging problems analyzed above from two aspects: data collection and algorithm design. In Tab.~\ref{table:datasets_compare}, we collect a large-scale FAS dataset based on surveillance scenes, namely SuHiFiMask. It has the following advantages: (a) \textbf{\emph{Rich surveillance scenes}.} It includes $40$ real surveillance scenes, such as movie theaters, security gates, and parking lots, which cover most face recognition scenes as much as possible. (b) \textbf{\emph{Realistic distribution of human faces and natural behavior.}}  It involves $101$ participants of different ages, and genders distribution participating in the data collection. These subjects perform natural behaviors in daily life. 
(c) \textbf{\emph{Rich spoofing Attacks.}} It has $232$ high-fidelity masks (\ie~resin, plaster, silicone, headgear, head mold), $200$ 2D attacks (\ie~posters, portraits, and screens), and $2$ adversarial attacks. %form an attack pool for data collection. 
(d) \textbf{\emph{Realistic lighting and diverse weather.}} We collect data under real outdoor scenes with different weather (\ie, sunny, snowy day) and light (\ie, day and night).
%Considering that real outdoor scenes are natural weather and light, we waited for four different kinds of weather to shoot and included both day and night scenes.

For the algorithm design, the Contrastive Quality-Invariance Learning network (CQIL) is proposed in Fig.~\ref{fig:CQIL}, which includes an Image Quality Variable (IQV) module and a two-stream framework consisting of a contrastive learning branch and a Separate Quality Network (SQN) branch. The IQV module is used to recover discriminative information related to FAS in the picture by super-resolution and deliver quality differences in contrast to the contrastive learning network backbone and SQN branch. The contrastive learning backbone~\cite{grill2020bootstrap} contains the online network and the target network. The online network continuously fits the target network during training, learning to approximate the same class with different quality distributions in the shared potential space. The SQN consists of a Quality-Invariance backbone network (CQI) (composed of a central differential convolution operator~\cite{yu2020searching}), a quality discriminator for separating quality, and the main classifier. CQI can effectively extract fine-grained features under environmental changes. The sample pairs generated by IQV are fed into CQI through adversarial learning, which allows CQI to focus on encoding features related to liveness while separating out the interference caused by quality. The main contributions of this paper are summarized below:
\begin{itemize}
	\setlength{\itemsep}{1.0pt}
	\item
	To the best of our knowledge, this is the first work to extend FAS to real surveillance scenes rather than mimicking low-resolution images and surveillance environments. We promote the development of this scenario through data collection and algorithm design.
	\item
	We collect a large-scale surveillance FAS dataset, SuHiFiMask, including $101$ participants of different ages, $232$ masks and $200$ 2D attacks. A total of $10,195$ videos were collected by 7 mainstream cameras in $40$ real scenes.
	\item
	We propose a novel Contrastive Quality-Invariance Learning (CQIL) network to enhance the detection of face attacks in surveillance. Among them, an Image Quality Variable (IQV) module is designed to recover the FAS information in images and construct sample pairs to simulate face quality differences in realistic surveillance. A contrastive learning branch to obtain features robust to quality changes. And a Separate Quality Network (SQN) branch based on adversarial learning is introduced to further guide the model to learn quality-independent liveness features.
	\item
	Extensive experiments are conducted on the SuHiFiMask and three other public datasets to demonstrate the challenges of the SuHiFiMask and the effectiveness of the proposed method.
\end{itemize}

\section{Related Work}

In this section, we review the current FAS works in constrained environments and some preliminary attempts in the surveillance scenes.

{\flushleft \textbf{FAS under constrained Environments.}}

Face spoofing (\eg, presentation attacks) is the typical physical attack to deceive the face recognition systems, where attackers present faces from spoof mediums, such as a photograph, screen, or mask, instead of a living human. According to the spoof mediums, we can roughly classify the existing attacks into 2D~\cite{Boulkenafet2017OULU,Liu2018Learning,zhang2020casia} and 3D attacks~\cite{nesli2013spoofing,liu2022contrastive}. Replay-Attack~\cite{chingovska2012effectiveness} and CASIA-FASD~\cite{zhang2012face} are early FAS datasets, commonly used as benchmark for domain generalization evaluation. The spoof medium of the former is an electronic screen, while the latter introduces additional paper mediums based on different resolutions. With the advancement of acquisition equipment in mobile phones, there are also some high-resolution datasets recorded by replaying face video with a smartphone, such as Replay-Mobile~\cite{costa2016replay}, OULU-NPU~\cite{Boulkenafet2017OULU}, and SiW~\cite{Liu2018Learning}. CelebA-Spoof~\cite{zhang2020celeba} introduces rich attribute annotation information, which can be used as an auxiliary task to improve the generalization of the model in various attacks. Recently, with the cost reduction of multi-spectral sensors and the popularity of use scenes, some new sensors have been introduced to provide more possibilities for FAS methods. Holger~\etal~\cite{steiner2016reliable} use multi-spectral short wave infrared (SWIR) imaging to ensure the authenticity of a face even in the presence of partial disguises and masks. Zhang~\etal~\cite{zhang2020casia} collect a CASIA-SURF dataset with $3$ modalities (\ie, RGB, Depth and NIR) using Intel RealSense SR300 camera, and propose a multi-modal multi-scale fusion method for FAS. Similarly, Liu~\etal~\cite{liu2021casia} introduce a CASIA-SURF CeFA dataset, covering $3$ ethnicities, $1,607$ subjects, and $23,538$ videos with $1280\times720$ resolution. As attack techniques are constantly upgraded, some new types of attacks have emerged, \eg, face mask~\cite{nesli2013spoofing,george2019biometric,liu2022contrastive}. Nesli~\etal~\cite{nesli2013spoofing} provide a 3DMAD which is recorded using the Microsoft Kinect sensor and consists of Depth and RGB modalities with 3D masks. George~\etal~\cite{george2019biometric} introduce a WMCA database with four channels, \eg, color, depth, near-infrared, and thermal, for face PAD which contains a wide variety of 2D and 3D presentation attacks, and propose MC-CNN method aiming to detect sophisticated attacks with multiple channels information. Heusch~\etal~\cite{heusch2020deep} collect an HQ-WMCA database, which can be viewed as an extension of the WMCA~\cite{george2019biometric} database via adding a new sensor acting in the shortwave infrared (SWIR) spectrum. A large-scale High-Fidelity Mask dataset, namely CASIA-SURF HiFiMask (briefly HiFiMask) was collected by Liu~\etal~\cite{liu2022contrastive}. Specifically, it consists of a total amount of $54,600$ videos which are recorded from $75$ subjects with $7$ kinds of sensors. Although the resolution and fidelity of these datasets are increasing high (\ie, resolution from $320\times240$~\cite{chingovska2012effectiveness} to $1,920\times1,080$~\cite{Liu2018Learning}, and spoofing types from print~\cite{zhang2012face} to mask~\cite{liu2022contrastive}), they are all oriented to FAS in a close constrained environment, ignoring the application requirements of remote surveillance scenes.

The essence of FAS is a defensive measure for face recognition systems and has been studied for over a decade. Early works were mainly based on color texture~\cite{chingovska2012effectiveness,boulkenafet2016faces} and motion analysis~\cite{Pan2007Eyeblink}. The former is based on the consideration that the fake face is different from the live face in texture details, such as color distortions, and specular highlights, due to the intervention of spoofing mediums. However, these algorithms are not accurate enough because of the use of handcrafted features, such as LBP~\cite{chingovska2012effectiveness}, HoG~\cite{schwartz2011face}, and SURF~\cite{boulkenafet2016face}. The latter analyzes the attack samples as static or non-rigid motion compared with live faces from the perspective of motion. Unfortunately, these methods become vulnerable if someone presents a replay attack or a print attack with cut eye/mouth regions. Instead of using pre-defined features such as LBP and HOG, CNN-based methods~\cite{yang2014learn,he2016deep} design a unified framework of feature extraction and classification in an end-to-end manner. However, they treat FAS as a binary classification task, and will highly depend on the liveness-unrelated cues, such as color distortion, shape deformation, or background information. Intuitively, the live faces in any scene have consistent face-like geometry. Inspired by this, some works~\cite{Liu2018Learning,wang2020deep,yu2020searching} leverage the physical-based depth information instead of binary classification loss as supervision, which are more faithful attack clues in any domain. Another works~\cite{Jourabloo2018Face,liu2022disentangling,stehouwer2020noise,liu2020disentangling,zhang2020face} treat FAS as a feature disentangled representation learning. Although these CNN-based methods achieve near-perfect performance under known attack clues, they still show poor generalization in the face of unknown attacks. To solve this problem, there are also some methods~\cite{wang2020cross,liu2021adaptive,liu2021dual,huang2022adaptive} that focus on improving the generalization of FAS in unknown domains. Examples are MADDG~\cite{shao2019multi}, SSDG~\cite{jia2020single}, are SSAN~\cite{wang2022domain}, which aim to learn a generalized feature space via adversarial training and triplet loss strategies. In the case of FMeta~\cite{shao2020regularized}, MT-FAS~\cite{qin2021meta}, D$^{2}$AM~\cite{chen2021generalizable}, and SDA~\cite{wang2021self}, they aim to find the generalized feature directions via meta-learning strategies.

{\flushleft \textbf{FAS in Surveillance Environments.}} The task of face recognition in surveillance has been widely concerned by researchers, including data collection and algorithm design. SCface~\cite{grgic2011scface} was the first face recognition dataset released to simulate research in surveillance scenes, which contains $4,160$ still images captured by five different quality cameras. The QMUL-Survface dataset~\cite{cheng2018surveillance} further complements the low-resolution face recognition dataset by collecting $463,507$ face images from $15,573$ different identities in the real world using surveillance cameras. Then, IJB-C~\cite{nada2018pushing} aims to improve the representation of the global population by adding a list of names containing specific occupations such as artists, public speakers, and journalists from different countries to the surveillance scenario. In addition, based on these datasets, face recognition algorithms for surveillance scenes have been in full swing. Li~\etal~\cite{8600370} introduce the adversarial generative networks and fully convolutional architectures to recognize ground-resolution faces in supervised discriminative learning. Considering the incompleteness of these datasets, Zhong~\etal~\cite{zhong2021sface} propose a sigmoid-constrained hypersphere loss (SFace) to reduce the intra-class distance of high-quality samples while preventing over-fitting label noise. Kim~\etal~\cite{Kim_2022_CVPR} propose an adaptive marginal function to adjust the importance of different samples by emphasizing the role of clean samples in classification. 

In the FAS community, Chen~\etal~\cite{chen2021dataset} explore the face anti-spoofing in surveillance scenes for the first time and proposed a dataset and benchmark. As for the dataset, they release the GREAT-FASD-S, which is first collected by two multi-modal cameras, and then processed into low-quality images. And for the method, they propose the DAM-SE module to select the most informative channels and recover the image with the nearest neighbor interpolation algorithm. Aravena~\etal~\cite{aravena2022impact} demonstrates that discarding a suitable percentage of low-quality samples can effectively improve the performance of the PAD algorithm. However, the nearest neighbor interpolation algorithm can not recover the original information by filling pixels with low-resolution images, and the method of directly discarding low-quality samples does not directly face the challenge of FAS in surveillance scenes.

%samples
\begin{figure}[ht]
	\centering
	\includegraphics[width=1.0\linewidth]{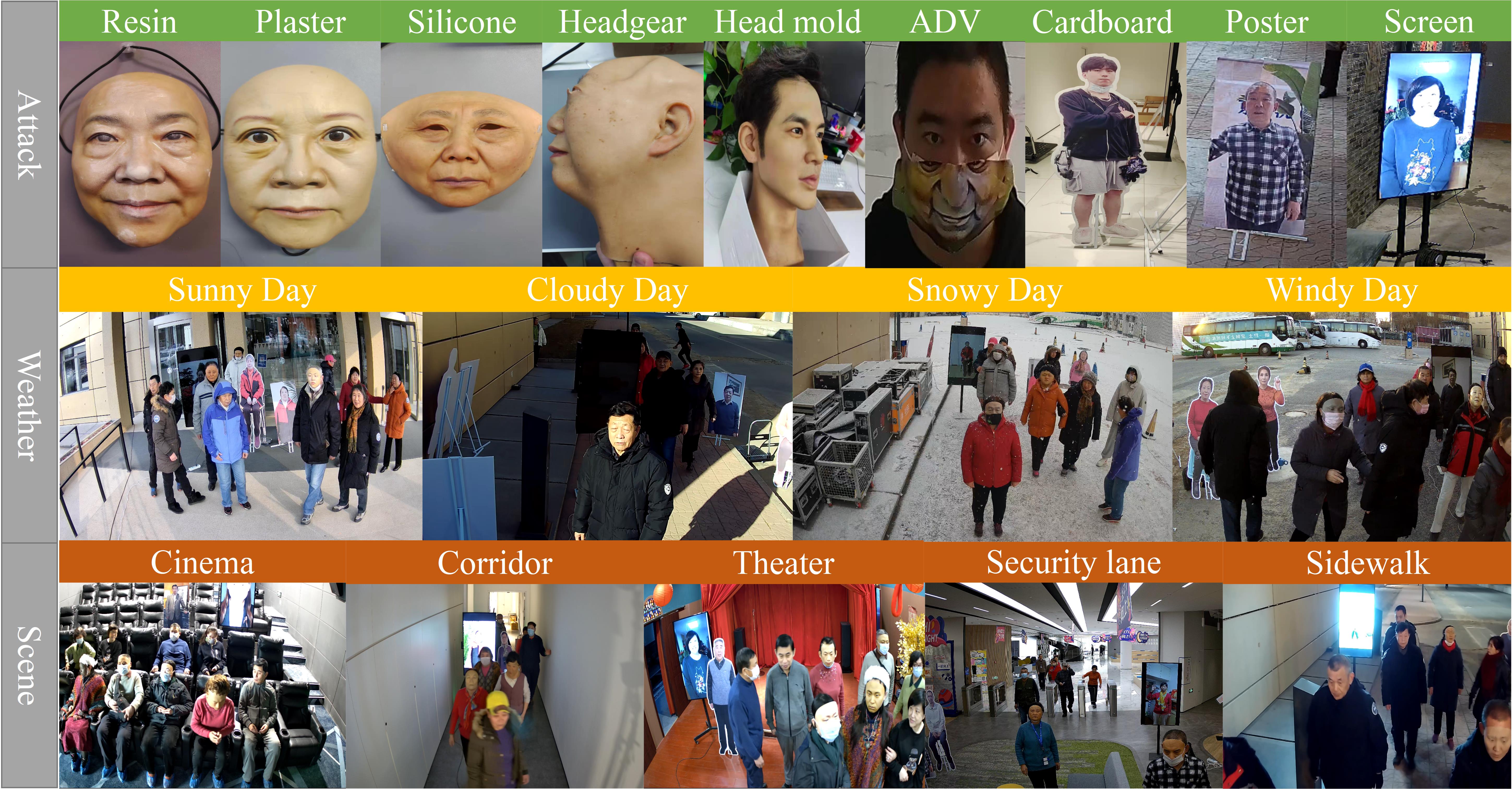}
	\vspace{-0.4cm}
	\caption{An overview of some characteristics of SuHiFiMask. From top to bottom: forms of attack, diverse weather, and surveillance scenes.}
	\label{fig:SuHiFiMask config}
\end{figure}
 
 \section{SuHiFiMask}

In order to fill the gap in the face anti-spoofing dataset of surveillance scenes and promote the research of related algorithms, we collected the SuHiFiMask dataset that has the following advantages over existing datasets:

\textbf{Advantage 1:} To the best of our knowledge, SuHiFiMask is the first dataset collected based on real surveillance scenes, rather than the low-quality datasets obtained by manual degradation, such as GREAT-FASD-S~\cite{chen2021dataset}. Compared to previous PAD datasets in controlled environments, the one we present inevitably introduces low-resolution face, pedestrian occlusion, changeable posture, motion blur, and other challenging situations, which greatly increases the challenge of FAS tasks. In addition, as shown in the third column of the Tab.~\ref{table:datasets_compare}, we define the dataset with the distance between the camera and the subject less than one meter as the short distance dataset, while the dataset with the distance between the camera and the subject greater than three meters is defined as the long-distance type. \textbf{Advantage 2:} SuHiFiMask considers the most comprehensive attack types, each of which contains diverse spoofing methods. As shown in Tab.~\ref{table:datasets_compare}, 2D image, video replay and 3D mask all appear in SuHiFiMask to evaluate the algorithm's perception of changes for paper color, screen moire and face structure in surveillance scenes. Different from the attack type under classical more constrained environments, as shown in Fig.~\ref{fig:SuHiFiMask config}, we introduce paper posters, humanoid stand-ups in 2D image, and headgear, head mold in the 3D mask to minimize the spoofing trace in the surveillance scenes. In order to effectively prevent criminals from hiding their identities through local occlusion during security inspection, we introduce two most effective adversarial attacks (ADV), instead of simply masking the face with paper classes~\cite{george2019biometric} and partial paper~\cite{liu2019deep}. \textbf{Advantage 3:} 

We designed $40$ common real-world surveillance scenes, including daily life scenes (\eg, cafes, cinemas, and theaters) and security check scenes (\eg, security check lanes and parking lots) for deploying face recognition systems. In fact, the rich natural behaviors in different surveillance scenes greatly increase the difficulty of PAD due to pedestrian occlusion and non-frontal views. \textbf{Advantage 4:} We collect data in four types of weather (\eg, Sunny, Windy, Cloudy and Snowy days) and natural lighting (\eg, Day and Night lights) to fully simulate the complex and changeable surveillance scenes. Different weather and light bring diverse image style information and image artifacts, which will put forward higher requirements for the generalization of PAD technology. 

Based on the above acquisition advantages, our SuHiFiMask contains $10,195$ videos from $101$ subjects of different age groups, which are collected by $7$ mainstream surveillance cameras and see Fig.1 in \emph{Appendix} for more details. In particular, as shown in the second and third rows of Fig.~\ref{fig:SuHiFiMask config}, SuHiFiMask is focused on surveillance scenes, and both real and fake attacks appear at the same time. 

\begin{figure}[ht]
	\centering
	\includegraphics[width=0.97\linewidth]{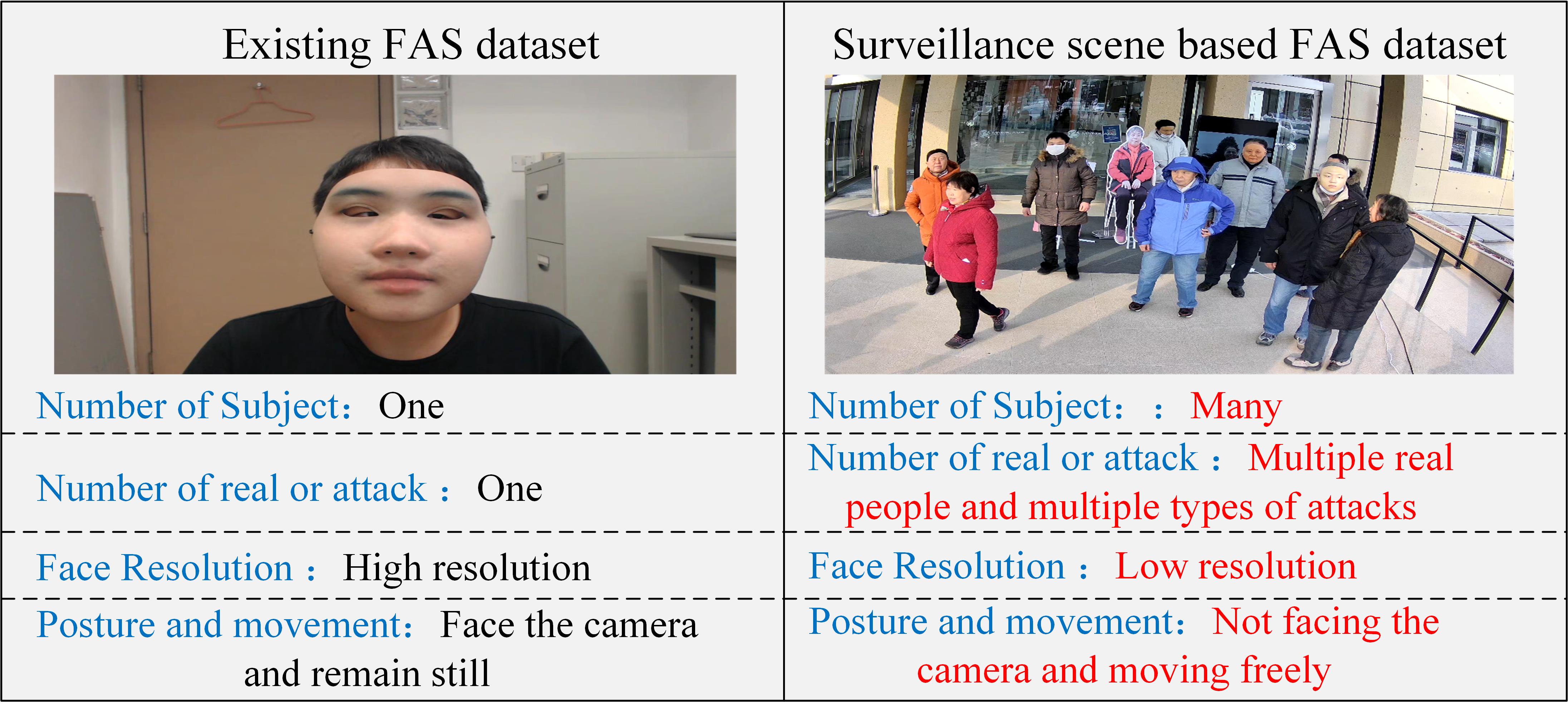}
	\caption{Comparison of the existing FAS dataset and the surveillance scene-based dataset. The left image is from MARsV2 and the right image is from the proposed SuHiFiMask.}
	\label{fig:contrast}
 \end{figure}

As shown in the Fig.~\ref{fig:contrast}, the existing FAS dataset of a video contains only one real person or one type of attack. The subject faces the camera and remains stationary during the shooting to ensure the clarity of the collected data. In contrast, videos based on a surveillance scene contain multiple real people and multiple types of attacks. Subjects are not required to face the camera and move randomly in the scene while filming. This leads to low-resolution of face images, pedestrian occlusions, non-frontal poses, and other disturbances that affect the stability and generalization of the algorithm. Thus, the surveillance scene-based FAS dataset poses a greater challenge than the existing FAS dataset.

\begin{table*}[ht]
\centering
\caption{Statistical information for each protocol of the proposed SuHiFiMask dataset. Note that 1, 2, 3 in the fourth column mean resin, silicone, and plaster. 4 represents headgear and head mold.}
\setlength{\tabcolsep}{5.4mm}
\scalebox{0.9}{
\begin{tabular}{|c|c|c|c|c|c|c|c|c|}
\hline
Pro.                 & Subset & \#Subject & Mask      & Quality score & \#Live & \#Mask & \#Other attack & \#All  \\ \hline
\multirow{3}{*}{1}   & Train  & 40      & 1\&2\&3\&4 & {[}0, 1{]}     & 118,520 & 60,715  & 22,333          & 201,568 \\ \cline{2-9} 
                     & Dev    & 10      & 1\&2\&3\&4 & {[}0, 1{]}     & 23,304  & 11,856  & 5663           & 40,823  \\ \cline{2-9} 
                     & Test   & 51      & 1\&2\&3\&4 & {[}0, 1{]}     & 69,878  & 42,569  & 19,743          & 132,190 \\ \hline
\multirow{3}{*}{2.1} & Train  & 101     & 1\&2\&3    & {[}0, 1{]}     & 100,990 & 40,454  & 0              & 141,444 \\ \cline{2-9} 
                     & Dev    & 101     & 1\&2\&3    & {[}0, 1{]}     & 20,521  & 19,608  & 0              & 40,129  \\ \cline{2-9} 
                     & Test   & 101     & 4          & {[}0, 1{]}     & 42,539  & 21,199  & 0              & 63,738  \\ \hline
\multirow{3}{*}{2.2} & Train  & 101     & 1\&2\&4    & {[}0, 1{]}     & 78,961 & 36,829  & 0              & 115,790 \\ \cline{2-9} 
                     & Dev    & 101     & 1\&2\&4    & {[}0, 1{]}     & 20,505  & 19,052  & 0              & 39,557  \\ \cline{2-9} 
                     & Test   & 101     & 3          & {[}0, 1{]}     & 42,521  & 28,366  & 0              & 70,887  \\ \hline
\multirow{3}{*}{2.3} & Train  & 101     & 1\&3\&4    & {[}0, 1{]}     & 77,952  & 28,994  & 0              & 106,946 \\ \cline{2-9} 
                     & Dev    & 101     & 1\&3\&4    & {[}0, 1{]}     & 20,594  & 17,714  & 0              & 38,308  \\ \cline{2-9} 
                     & Test   & 101     & 2          & {[}0, 1{]}     & 42,498  & 44,104  & 0              & 86,602  \\ \hline
\multirow{3}{*}{2.4} & Train  & 101     & 2\&3\&4    & {[}0, 1{]}     & 79,102 & 29,087  & 0              & 108,189 \\ \cline{2-9} 
                     & Dev    & 101     & 2\&3\&4    & {[}0, 1{]}     & 20,627  & 18,068  & 0              & 38,695  \\ \cline{2-9} 
                     & Test   & 101     & 1          & {[}0, 1{]}     & 42,513  & 42,887  & 0              & 85,400  \\ \hline
\multirow{3}{*}{3}   & Train  & 101     & 1\&2\&3\&4 & {[}0.4, 1{]}   & 64,276  & 35,898  & 58,889          & 159,063 \\ \cline{2-9} 
                     & Dev    & 101     & 1\&2\&3\&4 & {[}0.3, 0.4)   & 37,990  & 24,031  & 27,255          & 89,276  \\ \cline{2-9} 
                     & Test   & 101     & 1\&2\&3\&4 & {[}0, 0.3)     & 84,368  & 43,820  & 36,369          & 164,557 \\ \hline
\end{tabular}
}
\label{Table:Protocol}
\end{table*}

\subsection{Acquisition Details of SuHiFiMask.}

\textbf{Scenes and props.}
In order to cover real surveillance environments as much as possible, we carefully selected and rented $40$ real-world scenarios that include daily places, such as cafes, yoga studios, and movie theaters, as well as security checkpoints, such as security lanes, parking lots, and entrance/exit gates. We provide $232$ masks as the candidate pool for selection according to the scene requirements. Among them, some high-fidelity plaster and resin masks are from HiFiMask~\cite{liu2022contrastive}, and silicone material headgear and head mold masks are new additions to reduce the forgery traces exposed in the monitoring perspective, where the numbers of plaster masks, resin masks, silicone masks, headgear, and head mold were $93$, $93$, $23$, $10$, and $13$, respectively. In addition to mask attacks, we printed 2D images of $50$ subject in the form of humanoid upright cards and posters and provided video attacks by displaying images on a movable TV. In particular, in order to effectively prevent criminals from hiding their identity information during security checks in surveillance scenes, we crafted adversarial mask~\cite{zolfi2021adversarial} and adversarial hat~\cite{komkov2021advhat} that can induce face recognition systems to categorize the registered identity as unknown identity, aiming to increase the challenge to algorithm stability.

\textbf{Data collection and processing rules.} To ensure the quality and challenge of data, we implemented the following criteria before each shot: a) Device adjustment. We adjusted the positions and angles of each camera to ensure that the entire scene is captured. b) Sample balance. We arranged a consistent number of live and fake subjects to ensure sample balance in SuHiFiMask. c) Static 2D attacks. We deployed posters and humanoid upright-card and electronic screen-based photos with the same identity as the subjects at random locations. 

We also considered the following criteria during each shot: a) We designed specific movement routes for each subject to ensure adequate pedestrian occlusion, versatile posture and comprehensive perspective. b) We requested subjects in different scenes to perform scene-related behaviors, such as eating and chatting in daily scenes, and self-service check-in security check scenes. 

After data collection, we performed the following pre-processing: a) Face detection. We used RetinaFace~\cite{deng2020retinaface} to detect the face in each frame of the original video, and discarded those frames where the face could not be detected. b) Face tracking. We use face similarity to track the position of faces in consecutive frames and name each of the different face tracking boxes. c) Video sampling. We sample each video in 10-frame intervals and store the cropped face image in the corresponding face tracking box folder. d) Dataset naming rules. We named the folder of this video according to the following rule: $Group\_Scene\_Camera\_Epoch\_Time$.

\textbf{Ethical and legal considerations.} 
% and dispatched multiple security personnel to maintain order at the shoot
Since we collected our filming scenes from real-world environments, we have a responsibility to maintain the public environment and protect pedestrian safety. We commissioned two companies to legally authorize the scenes for data collection. SuHiFiMask is a dataset consisting of videos taken from subjects of different age groups, and although this is not a subject explicitly modeled for human behavior, the relevant challenge factors are related to humans. Based on the consideration of the protection of human rights and legal interests, our collection process follows a strictly ethical procedure. We commission a data acquisition company to develop strict standards and obtain the signature authorization of all human subjects. The collected images and videos will be used to develop, train and optimize face anti-spoofing technologies to the extent permitted by Chinese laws. The dataset is balanced in terms of gender and age, that is, there is no hazard in terms of ethics.

\subsection{Evaluation protocol and Statistics.}

We define three protocols for SuHiFiMask to fully evaluate the performance in surveillance environments: Protocol 1-ID, Protocol 2-Mask, and Protocol 3-quality.

\textbf{Protocol 1-ID.} Protocol 1 aims to evaluate the comprehensive performance of the algorithm being migrated to long-distance surveillance scenes. Compared with the classical constrained environment datasets, protocol 1 includes various unique factors in surveillance scenes, such as low resolution, pedestrian occlusion, changeable posture, motion blur, and other complex weather, which pose greater challenges to algorithm design. As shown in Tab.~\ref{Table:Protocol}, we divide the training set, development set, and testing set according to the identity information, including $40$, $10$, and $50$ subjects, respectively.

\textbf{Protocol 2-Mask.} Protocol 2 evaluates the generalization of the algorithm for the `unseen' 3D facial mask type. The diversity and unpredictability of mask materials are important characteristics of spoofing means and are easily interfered with by other liveness-unrelated factors. Thus, the generalization to mask materials is an important evaluation index. In this work, we divide protocol 2 into four sub-protocols by using the `leave-one-type-out testing' method, in which one unknown 3D mask material is divided into the testing set for each sub-protocol. As shown in Tab.~\ref{Table:Protocol}, `$1$', `$2$', `$3$', and `$4$' in the fourth column indicate that the 3D mask material is headgear/head mold, resin, silicone, and plaster, respectively.

 \begin{figure}[ht]
	\centering
	\includegraphics[width=1.0\linewidth]{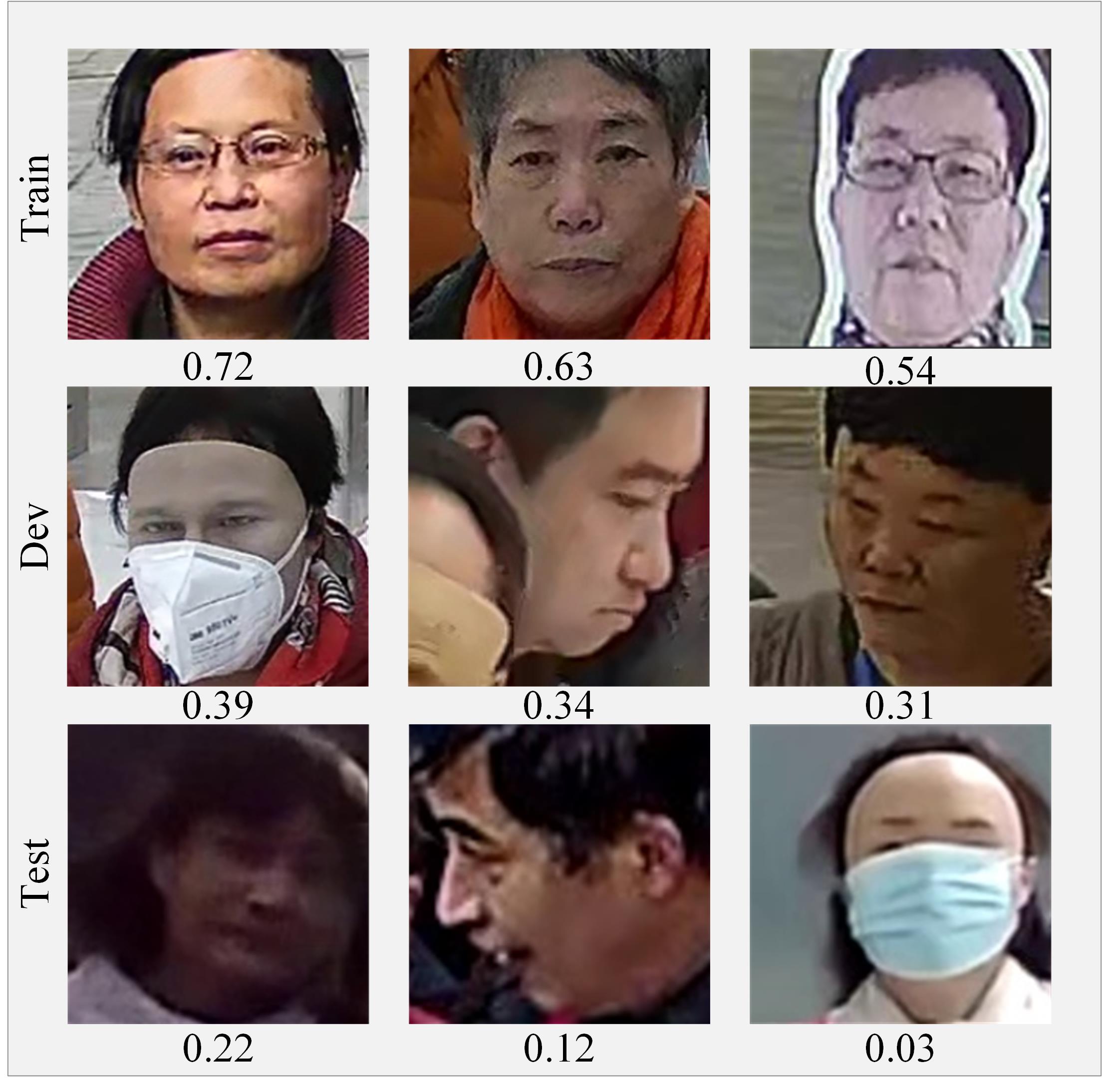}
	\caption{Some samples from protocol 3. The number below the image represents the quality score of the faces. Samples with different score intervals are proportionally categorized as the training set, development set, and testing set.}
	\label{fig:quality}
 \end{figure}

\textbf{Protocol 3-Quality.} Protocol 3 evaluates the robustness of the algorithm to image quality degradation. Variable quality and disturbances are factors that affect the stability of the algorithm. Therefore, the robustness of the algorithm to quality degradation is an important metric to be evaluated. In this work, as shown in Fig.~\ref{fig:quality}, we use the SER-FIQ~\cite{terhorst2020ser} algorithm to calculate the image quality score which ranges from 0 to 1. As shown in the fifth column of Tab.~\ref{Table:Protocol}, we assign images with scores [0.4, 1] as the training set, scores [0.3, 0.4) as the development set, and scores [0, 0.3) as the testing set.

\section{Methodology}
\begin{figure*}[ht]
	\centering
	\includegraphics[width=0.95\linewidth,height=0.43\textwidth]{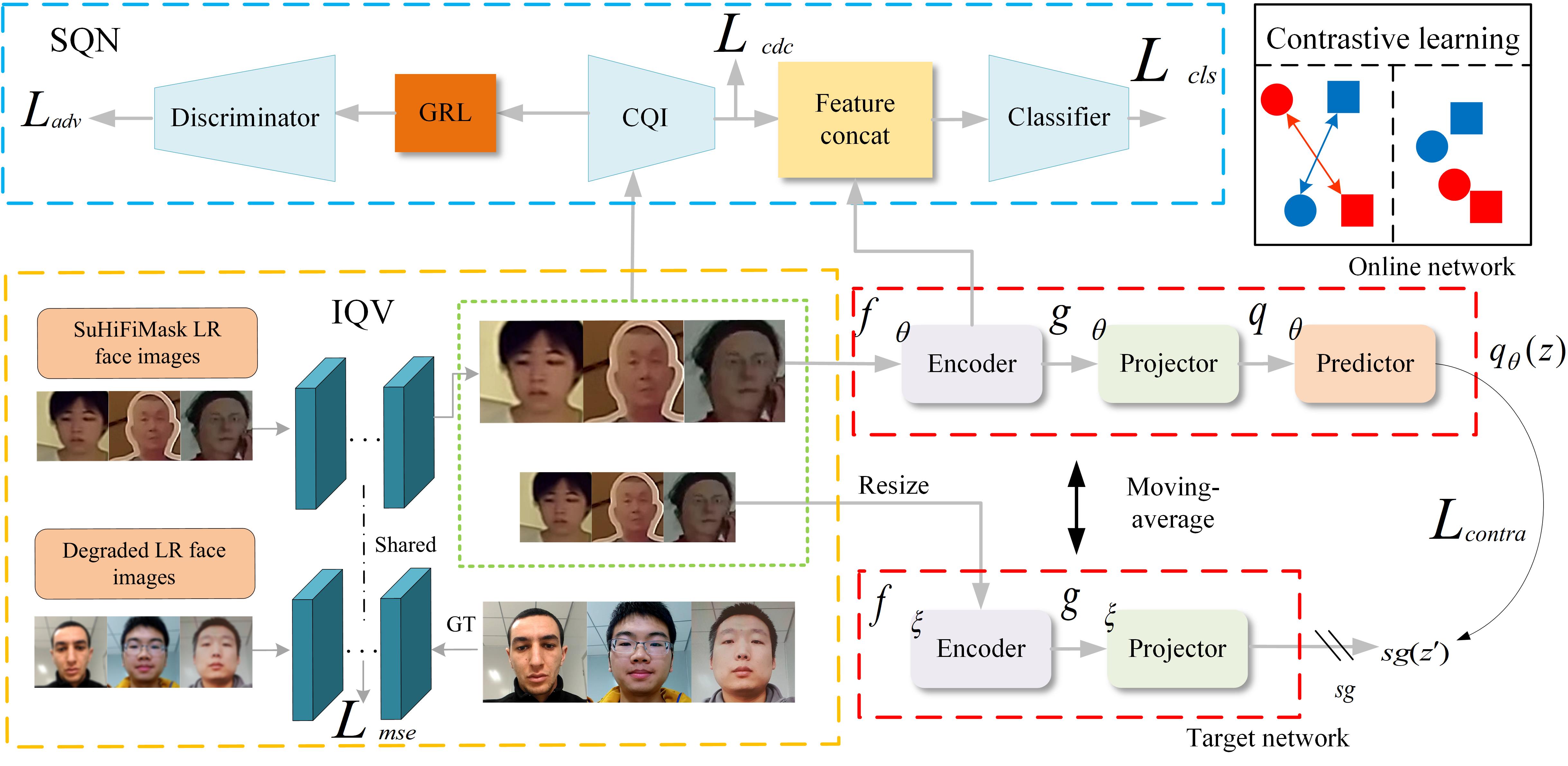}
	\vspace{-0.2cm}
	\caption{Contrastive Quality-Invariance Learning (\text{CQIL}) network. IQV recovers information from the images and constructs sample pairs of different qualities. Sample pairs are sent to the contrastive learning branch and the SQN branch. The contrastive learning branch consists of an online network (encoder, projector, predictor) and a target network (encoder, projector). Above the image the SQN branch is shown, which contains the discriminator, CQI, GRL, and the main classifier.}
	\label{fig:CQIL}
\end{figure*}

In this section, we present a Contrastive Quality-Invariance Learning (CQIL) network for FAS tasks based on long-distance surveillance scenes. As shown in Fig.~\ref{fig:CQIL}, CQIL contains an Image Quality Variable module (IQV) and a dual-stream framework with a contrastive learning branch and a Separate Quality Network (SQN) branch. IQV processes low-quality images into high-quality images by super-resolution and sends them to the contrastive learning branch and the SQN branch. The contrastive learning branch trains the network by using high-quality and low-quality images as input to the online network and the target network, respectively. The SQN branch makes the features extracted by the encoder independent of quality by adversarial learning. In addition, CQI uses high-quality images after super-resolution as input to extract richer discriminative features.

{\flushleft \textbf{Image Quality Variable Module (IQV).}}
In contrast to the classical constrained environment, the difficulty of the FAS task based on surveillance scenes is the low resolution and variable quality of the images, which leads to insufficient information contained in the images and severely interferes with the extraction of robust features. To solve this problem, a possible solution is to increase the resolution of the image and extract robust invariant features. Inspired by CSRI~\cite{cheng2018low}, we introduce the Image Quality Variable (IQV) module to improve the image resolution of SuHiFiMask and recover information relevant to the FAS task. In addition, IQV tags the images processed by the SR network with label 0 and the original images with label 1. Then IQV sends them to the contrastive learning branch and the SQN branch. Since SuHiFiMask is the first unconstrained PAD dataset, there is no high-quality image as ground truth to optimize the super-resolution network. Thus, we use the existing high-definition PAD dataset to train the super-resolution network. As shown in Fig.~\ref{fig:CQIL}, this process can be expressed as follows: 1) We degrade the high-fidelity dataset OULU-NPU~\cite{Boulkenafet2017OULU} into a low-quality dataset using pre-processing methods such as interpolation and gaussian blurring. 2) We feed degraded low-resolution images into an SR network and use its original data for supervision to train the SR network. 3) We use the SR network with shared parameters to process SuHiFiMask's images into high-quality images. Unlike the standalone super-resolution tasks, we combine the SR tasks with the FAS tasks by integrating the IQV module into the framework with the following two advantages below: 
\begin{itemize}
    \item Training the FAS network with SR network-boosted resolution images can improve the performance of the FAS network.
    \item The improved performance of other networks in CQIL can better guide the SR network to recover information related to the FAS task in the image. 
\end{itemize} 
Finally, MSE loss is used to constrain the super-resolution network:
\begin{equation}
\mathcal{L}_{mse}=\frac{1}{\mathrm{n}} \sum\left(\hat{\mathrm{y}}_{\mathrm{i}}-\mathrm{y}_{\mathrm{i}}\right)^{2}
 \label{eq:MSE}
\end{equation}
where $n$ represents the number of pixels in the image, $\hat{\mathrm{y}}_{\mathrm{i}}$, $\mathrm{y}_{\mathrm{i}}$ denote the pixel value of the image after super-resolution and the pixel value of ground truth respectively.

{\flushleft \textbf{Contrastive Learning Branch.}}
To improve the robustness of FAS networks in a quality-variant surveillance environment, we propose a branch based on contrastive learning. Inspired by the BYOL~\cite{grill2020bootstrap}, this branch obtains robustness to quality variations by fitting the distribution of potential features for different quality pictures. Specifically, during the training process, due to the constraints of Eq.~\ref{eq:contrast} and Eq.~\ref{eq:contrast2}, the online network will gradually fit the target network by closing the same class in the potential feature space for pairs of images of different quality, which makes it to obtain a powerful feature representation while ignoring the negative impact from different quality distributions.
\begin{equation}
\mathcal{L}_{\theta, \xi} \triangleq\left\|\overline{q_{\theta}}\left(z_{\theta}\right)-\bar{z}_{\xi}^{\prime}\right\|_{2}^{2}=2-2 \cdot \frac{\left\langle q_{\theta}\left(z_{\theta}\right), z_{\xi}^{\prime}\right\rangle}{\left\|q_{\theta}\left(z_{\theta}\right)\right\|_{2} \cdot\left\|z_{\xi}^{\prime}\right\|_{2}} 
\label{eq:contrast}
\end{equation}
\begin{equation}
\mathcal{L}_{contra}=\mathcal{L}_{\theta, \xi}+\widetilde{\mathcal{L}}_{\theta, \xi}
\label{eq:contrast2}
\end{equation}
where $q_{\theta}\left(z_{\theta}\right)$ is the prediction of the online network output and $z_{\xi}^{\prime}$ is the projection of the target network output, then we use $\ell_{2} \text {-normalize }$ to turn $q_{\theta}\left(z_{\theta}\right)$ and $z_{\xi}^{\prime}$ into $\overline{q_{\theta}}\left(z_{\theta}\right)$ and $\bar{z}_{\xi}^{\prime}$. In addition, $\widetilde{\mathcal{L}}_{\theta, \xi}$ is the result of $\mathcal{L}_{\theta, \xi}$ symmetrization.

As shown in Fig.~\ref{fig:CQIL}, image pairs of different quality generated by IQV are sent to the online and target networks. The online network is composed of an encoder network (Interchangeable backbone networks), a projector (Projection of extracted features into the latent space), and a predictor (with the same multi-layer perceptron structure). Similarly, the target network has an encoder and a projector with different weights from the online network. Unlike the weight update of the online network, the parameters of the target network are not updated in gradient descent~\cite{grill2020bootstrap}, and the process can be expressed as follows:
\begin{equation}
{\xi} {\gets} {\tau} {\xi}+ (1- {\tau}) {\theta}
\label{eq:update}
\end{equation}
The parameters ${\xi}$ and ${\theta}$ represent the parameters to be updated for the target network and the online network, respectively. The parameters ${\theta}$ of the online network are updated by the optimization of the loss function, the parameters ${\xi}$ of the target network are updated by perceiving an exponential moving-average~\cite{he2020momentum} of the online parameters and we perform the moving-average after each step by target decay rate ${\tau}$.

{\flushleft \textbf{Separate Quality Network (SQN).}}
For FAS data in surveillance scenes, which contains many variations (\eg, environment, light, weather), we need operators that are more robust to variations to describe the required fine-grained information. Inspired by central differential convolution (CDC)~\cite{yu2020searching}, we use CDC to form a quality-independent backbone network (CQI) in the second branch, exploiting its powerful representation ability to extract fine-grained features under environmental variations. In addition, we use cross-entropy loss as a supervision of CQI, so that this network can capture the cues related to liveness more robustly. 

The sample pairs generated by the IQV module have the following characteristics: 1) Both the super-resolution network processed images and the original images contain the object of the face (live or attack) in the center of their images, so even samples with very different quality share the same semantic feature space. 2) Although the quality of each image is different, they all contain discriminative information. Therefore, we make the discriminative features extracted by CQI independent of quality by adversarial learning. Specifically, we use the adversarial loss to optimize the backbone network CQI. And the gradient reversal layer (GRL)~\cite{ganin2015unsupervised} allows the parameters of the quality discriminator to be optimized in the reverse direction. This process can be formulated as follows:
 \begin{equation}
 \begin{aligned}
&{\min_D}{\max_C}{\mathcal{L}_{adv}}(C,D)=\\
&-E_{(x.y){\sim}(X,Y_Q)}{\sum_{i=1}^{N}}1[i=y]{{\log}D(C(x))}
\end{aligned}
\label{eq:adv}
\end{equation}
where ${Y_Q}$ is the set of quality labels, $N$ is the number of images of different quality, $C$ stands for the CQI network backbone where we extracted the liveness-related information, and $D$ represents the quality discriminator. Finally, we concatenate the features extracted by CQI with those extracted by the contrastive learning branch and input them to the classifier for classification.

\begin{algorithm}[ht]
\caption{Contrastive Quality-Invariance Learning (CQIL)} %算法的名字
\hspace*{0.02in} {\bf Input:} %算法的输入, \hspace*{0.02in}用来控制位置,同时利用 \\ 进行换行
image set $X$, label set $Y$, HD image set $H$.\\     
% \hspace*{0.02in} {\bf Output:} %算法的结果输出
% output result
\begin{algorithmic}[1]    % \State 后写一般语句
\State Initialize: encoder ${f}_\theta$, projector ${g}_\theta$, predictor ${q}_\theta$ 
\State Initialize: encoder ${f}_\xi^{\prime}$, projector ${g}_\xi^{\prime}$
\State Initialize: network $n$ of IQV, encoder $c$ of CQI 
\While{not end of training} % For 语句,需要和EndFor对应
   \State sample batch $\mathcal{A}\leftarrow\left\{x_{i} \sim X\right\}_{i=1}^{N}$, $\mathcal{B}\leftarrow\left\{y_{i} \sim Y\right\}_{i=1}^{N}$
   \State sample batch $\mathcal{H}\leftarrow\left\{h_{i} \sim H\right\}_{i=1}^{N}$   
   \For{$x_{i} \in \mathcal{A}$} 
       \State  $l_{i} \leftarrow h_i$      \Comment{Degradation into lq images}
       \State \textbf{compute} {$\mathcal{L}_{mse}$}, see Eq.~\ref{eq:MSE}
       \State $x_{i1}\leftarrow n(x_i), x_{i2}\leftarrow x_i$  \Comment{Generate image pairs}
       \State $y_{i1}, y_{i2} \leftarrow x_{i1}, x_{i2}$\Comment{Generate quality labels}
       \State $z_{1} \leftarrow g_{\theta}\left(f_{\theta}\left(x_{i1}\right)\right)$, $z_{2} \leftarrow g_{\theta}\left(f_{\theta}\left(x_{i2}\right)\right)$
       \State $z_{1}^{\prime}\leftarrow g_{\theta}\left(f_{\theta}\left(x_{i2}\right)\right)$, $z_{2}{\prime} \leftarrow g_{\theta}\left(f_{\theta}\left(x_{i1}\right)\right)$
       \State \textbf{compute} {$\mathcal{L}_{contra}$}, see Eq.~\ref{eq:contrast}, Eq.~\ref{eq:contrast2}
       \State $\mathcal{X}\leftarrow [x_{i1},x_{i2}]$, $\mathcal{Y}\leftarrow [y_{i1},y_{i2}]$, $\mathcal{Z}\leftarrow c(\mathcal{X})$
       \State \textbf{compute} {$\mathcal{L}_{adv}$}, by Eq.~\ref{eq:adv}
       \State $z_3\leftarrow c({x_i})$, $z_4\leftarrow f_{\theta}(x_i)$
       \State \textbf{compute} {$\mathcal{L}_{cls}$}, {$\mathcal{L}_{cdc}$}, {$\mathcal{L}_{total}$}, by Eq.~\ref{eq:total}
   \EndFor
\EndWhile
\State $\Delta \theta= { backward }\left(\mathcal{L}_{total }\right)$
\State $\theta \leftarrow \theta {-} {learningrate} \cdot \Delta \theta$
\State \textbf{update} $\xi$ by Eq.~\ref{eq:update}
\State \textbf{update} network $n$, encoder $c$
\end{algorithmic}
\end{algorithm}

{\flushleft \textbf{Overall Loss.}}
As mentioned, CQI is used to extract quality-independent discriminative features, and these features are concatenated with the robust features extracted from the contrastive learning branch and fed to the main classifier. Therefore, the cross-entropy loss ${\mathcal{L}_{cdc}}$ and ${\mathcal{L}_{cls}}$ is well constrained for both CQI and the main classifier. In summary, the overall loss function ${\mathcal{L}_{total }}$ for stable and reliable training can be formulated as follows:
 \begin{equation}
 \begin{aligned}
\mathcal{L}_{ {total }}=\lambda_{1}\cdot\mathcal{L}_{{cls}}+\lambda_{2}\cdot\mathcal{L}_{ {contrast}}+\lambda_{3}\cdot\mathcal{L}_{{adv}}\\+\lambda_{4}\cdot\mathcal{L}_{ {cdc}} + \lambda_{5}\cdot\mathcal{L}_{ {mse}}
\end{aligned}
\label{eq:total}
\end{equation}
where${\lambda_{1}}$, ${\lambda_{2}}$, ${\lambda_{3}}$, ${\lambda_{4}}$ and ${\lambda_{5}}$ are five hyper-parameters to balance the proportion of the different loss functions.

\section{Experiments}
\subsection{Experiments Settings}
\textbf{Dataset and Protocols.} In experiments, a total of five datasets were used: OULU-NPU~\cite{Boulkenafet2017OULU}, CASIA-MFSD~\cite{zhang2012face}, RepalyAttack~\cite{chingovska2012effectiveness}, MARsV2~\cite{liu20163d} and the SuHiFiMask dataset. First, we conducted ablation experiments on three protocols of the proposed SuHiFiMask to demonstrate the effectiveness of each component of the proposed CQIL. Second, we present the respective baselines for the different protocols for the proposed dataset. Finally, we design several different cross-testing experiments to demonstrate the importance of the proposed dataset and the effectiveness of the method.

\textbf{Training Setting.}
Our proposed method is implemented with Pytorch. In the training stage, models are trained with Adam optimizer and the initial learning rate is $2e-4$. The batch size is set to $6$ for CQIL. The epoch of the intra-testing is set to $10$, and the lr decreases by $0.2$ times per epoch. The epoch of the inter-testing is $300$, and lr decreases by $0.2$ times per $50$ epochs. ${\lambda_{1}}$, ${\lambda_{2}}$, ${\lambda_{3}}$, ${\lambda_{4}}$ and ${\lambda_{5}}$ are set to $2$, $1.5$, $0.5$, $1.5$, $0.5$ respectively.
% In the intra-testing experiment, ${\lambda_{2}}$, ${\lambda_{4}}$ is set to $1$, $2$ on protocol 1, $1.8$, $1.5$ on protocol 2.1, $1.5$, $1.5$ on protocols 2.2, 2.3, 2.4, and $1.7$, $1.4$ on protocol 3. In inter-testing experiments, ${\lambda_{2}}$, ${\lambda_{4}}$ are set to $2$, $1$ when tested on MARsV2 and MARsV2-5×5, and ${\lambda_{2}}$, ${\lambda_{4}}$ are set to $1.5$, $1.5$ when tested on MARsV2-3×3.

\textbf{Performance Metrics and Implementation Details.} We accept the Attack Presentation Classification Error Rate (APCER), Bonafide Presentation Classification Error Rate (BPCER), and ACER~\cite{ACER} as the evaluation metrics in our experiments. The ACER on each testing set is determined by the threshold value of the performance on the development set. In cross-testing experiments, we use Half Total Error Rate (HTER)~\cite{bengio2004statistical} and Area Under Curve (AUC) as evaluation metrics. We use the ResNet18~\cite{he2016deep},  ViT~\cite{dosovitskiy2020image}, and the CDCN~\cite{yu2020searching} network as the backbone, and report their results in experiments.

\subsection{Ablation Study.}
Here we conduct ablation experiments to verify the contribution
of each module of the proposed CQIL on the three protocols of the SuHiFiMask dataset.
\begin{table}[ht]
\centering
\caption{The ablation study of different Components. The evaluation metric is ACER (\%).}
\scalebox{1.3}{
\begin{tabular}{|c|c|c|c|}
\hline
Method        & Prot.1                     & Prot.2                             & Prot.3                     \\ \hline
ResNet18      & 12.58                      & 16.55$\pm$51.71                      & 17.64                      \\
CQIL-Model-1 & \multicolumn{1}{c|}{11.97} & \multicolumn{1}{c|}{16.01$\pm$50.23} & \multicolumn{1}{c|}{17.45} \\
CQIL-Model-2 & 11.75                      & 15.67$\pm$48.12                      & 16.54                      \\
CQIL-Model-3 & 10.90                      & 15.14$\pm$46.66                      & 16.13                      \\
\textbf{CQIL-Model-4} & \textbf{10.69}             & \textbf{14.90$\pm$45.92}          & \textbf{15.98}                      \\ \hline
\end{tabular}
} 
\label{table:abla}
\end{table}

\textbf{Advantage of the proposed architecture.} We compare four architectures with ResNet18 to demonstrate the advantages of each module of the proposed method. The CQIL-model-1 is a contrastive learning network with ResNet18 as its backbone. Since the training of the contrastive learning network requires the output of the IQV module, we use images processed by cubic interpolation and nearest-neighbor interpolation to mimic samples of different quality to eliminate the impact of the IQV module on performance. In addition, we additionally supervise the training of the online encoder using cross-entropy loss. In the testing phase, we use the features extracted by the online encoder for classification. As shown in Tab.~\ref{table:abla}, CQIL-model-1 has a significant improvement in performance on all three protocols compared to ResNet18, which demonstrates that the contrastive learning branch using quality change as a contrast improves the robustness of the network in surveillance scenes.

\textbf{Advantage of SQN branch.} Our proposed SQN branch takes sample pairs of different qualities generated by the IQV module as input and lets the discriminative features extracted by the encoder CQI be independent of the quality by adversarial learning. CQIL-model-2 extends the SQN branch on the basis of CQIL-model-1. In the testing phase, we concatenate the features extracted by the contrastive learning branch with the features extracted from CQI in the SQN branch for classification. As shown in Tab.~\ref{table:abla}, the performance of CQIL-model-2 is significantly improved on all three protocols, and the performance improvement is especially obvious in protocol 3, which verifies that SQN trained with samples of different quality have the ability to extract discriminative features independent of quality.

\textbf{Advantage of IQV module.} CQIL-model-3 extends the complete IQV module based on CQIL-model-2 but uses low-quality original images to train the CQI encoder. CQIL-model-4 extends CQIL-model-3 by training CQI encoders using high-quality images generated by SR networks. The improved performance of CQIL-model-3 in Tab.~\ref{table:abla} demonstrates that the sample pairs constructed by IQV more closely match the quality variation in the surveillance scene and IQV can effectively improve the performance of the SQN branch and contrastive learning branch. The improved performance of CQIL-Model-4 further validates the two advantages of IQV modules: 1) The SR network processed images can be used for CQI encoder training, thus improving the performance of the FAS task. 2) The performance-improved FAS network can better guide the SR network to recover the discriminative information of the images.
\subsection{Intra-Testing.}
Here, we conduct experiments on three different protocols of SuHiFiMask, showing that SuHiFiMask poses a challenge to existing FAS studies while also testing the performance of our proposed CQIL method in different data distributions.
\begin{table}[ht]
\centering
\caption{The results of intra-testing on three protocols of SuHiFiMask.}
\scalebox{0.95}{
\begin{tabular}{|c|c|c|c|c|}
\hline
Prot.                & Method     & APCER\%                       & BPCER\%                      & ACER\%                       \\ \hline
\multirow{4}{*}{1} & ResNet18   & 13.59                         & 11.57                        & 12.58                        \\
                   & ViT        & 13.45                         & 9.89                         & 11.67                        \\
                   & CDCN       & 20.46                         & 18.95                        & 20.41                       \\
                   & \textbf{CQIL (ours)} & \textbf{11.09}                & \textbf{10.29}               & \textbf{10.69}               \\ \hline
\multirow{4}{*}{2} & ResNet18   & 20.46$\pm$184.60  & 12.74$\pm$1.43 & 16.60$\pm$51.05 \\
                   & ViT        & 19.56$\pm$181.71 & 12.25$\pm$0.42 & 15.89$\pm$45.01 \\
                   & CDCN       & 24.88$\pm$55.77   & 24.44$\pm$12.51 & 24.66$\pm$16.46  \\
                   & \textbf{CQIL (ours)} &\textbf{ 18.83${\pm}$169.37} & \textbf{10.88${\pm}$0.34}   & \textbf{14.86${\pm}$46.04} \\ \hline
\multirow{4}{*}{3} & ResNet18   & 21.04                         & 13.64                        & 17.64                      \\
                   & ViT        & 19.61                         & 13.95                        & 16.78                       \\
                   & CDCN       & 28.70                         & 25.89                        & 27.30                        \\
                   & \textbf{CQIL (ours)} & \textbf{19.14}                & \textbf{12.82}               & \textbf{15.98}            \\ \hline
\end{tabular}
}
\label{table:Protocol1}
\end{table}

\textbf{Experiments on Protocol 1-ID.} 
In protocol 1, the data distribution is similar for different sets. The training set, development set, and testing set contain all attack types, and also contain data for all quality scores. The protocol is appropriate to evaluate the performance of the FAS algorithm in long-distance surveillance scenes. As shown in Tab.~\ref{table:Protocol1}, the proposed CQIL ranks first for three performance metrics (11.09\%, 10.29\%, 10.69\%, respectively) compared to the generic network backbone ResNet, ViT, and the FAS task network CDCN with robust feature representation on the Protocol 1, showing that the proposed method performs well in the FAS task based on surveillance scene with low resolution and many interferences.

\textbf{Experiments on Protocol 2-Mask.} 
We verify the algorithm's ability to discriminate between different types of masks by protocol 2. As shown in Tab.~\ref{table:Protocol2}, our proposed CQIL achieved good results except for the APCER on protocol 2.1 and protocol 2.2 which was not the highest performance, which proves that our method can extract discriminative features in low-quality mask images. It is worth mentioning that the testing set of protocol 2.1 is composed of headgear and head mold. These two types of masks are very similar to the human head structure, so the algorithm can no longer use features such as mask contours as a basis for prediction. Thus, the performance of CQIL on protocol 2.1 demonstrates the importance of CQI encoders that can extract fine-grained features.

\begin{table}[ht]
\centering
\caption{The results of intra-testing on four sub-protocols of SuHiFiMask protocol 2.}
\scalebox{1.1}{
\begin{tabular}{|c|c|c|c|c|}
\hline
Prot.                 & Method     & APCER\%        & BPCER\%                            & ACER\%         \\ \hline
\multirow{4}{*}{2.1} & ResNet18   & 43.33          & \multicolumn{1}{c|}{13.96}         & 28.65          \\
                     & ViT        & 42.54          & \multicolumn{1}{c|}{12.05}         & 27.29          \\
                     & CDCN       & \textbf{35.47} & \multicolumn{1}{c|}{20.69}         & 28.08                \\
                     & \textbf{CQIL (Ours)} & 41.18          & \multicolumn{1}{c|}{\textbf{11.81}} & \textbf{26.49} \\ \hline
\multirow{4}{*}{2.2} & ResNet18   & \textbf{8.50}  & \multicolumn{1}{c|}{11.58}          & 10.04          \\
                     & ViT        & 8.56           & \multicolumn{1}{c|}{11.40}          & 9.98          \\
                     & CDCN       & 14.72          & \multicolumn{1}{c|}{21.41}         & 18.06          \\
                     & \textbf{CQIL (Ours)} & 8.88 & \multicolumn{1}{c|}{\textbf{10.26}} & \textbf{9.57} \\ \hline
\multirow{4}{*}{2.3} & ResNet18   & 17.52          & \multicolumn{1}{c|}{11.51}         & 14.52          \\
                     & ViT        & 13.70          & \multicolumn{1}{c|}{12.33}         & 13.02          \\
                     & CDCN       & 26.62          & \multicolumn{1}{c|}{29.20}         & 27.91          \\
                     & \textbf{CQIL (Ours)} & \textbf{13.61}        & \multicolumn{1}{c|}{\textbf{10.92}} & \textbf{12.27} \\ \hline
\multirow{4}{*}{2.4} & ResNet18   & 12.48           & \multicolumn{1}{c|}{13.90}        & 13.19          \\
                     & ViT        & 13.33          & \multicolumn{1}{c|}{13.20}         & 13.26         \\
                     & CDCN       & 22.72          & \multicolumn{1}{c|}{26.47}         & 24.59          \\
                     & \textbf{CQIL (Ours)} & \textbf{11.64}          & \multicolumn{1}{c|}{\textbf{10.54}} & \textbf{11.09} \\ \hline
\end{tabular}
}
\label{table:Protocol2}
\end{table}

\begin{table}[ht]
\centering
\caption{The results of cross-dataset testing for CASIA-MFSD, Replay-Attack and SuHiFiMask. The evaluation metric is HTER($\%$).}
\scalebox{0.98}{

\begin{tabular}{|cc|cc|cc|}
\hline
\multicolumn{1}{|c|}{\multirow{2}{*}{Method}} & Train & \multicolumn{2}{c|}{CASIA-MFSD}                                                                                                        & \multicolumn{2}{c|}{ReplayAttack}                                                                                                   \\ \cline{2-6} 
\multicolumn{1}{|c|}{}                        & Test  & \multicolumn{1}{c|}{\begin{tabular}[c]{@{}c@{}}Replay-\\ Attack\end{tabular}} & \begin{tabular}[c]{@{}c@{}}\textbf{SuHiFi-}\\ \textbf{Mask (Ours)}\end{tabular} & \multicolumn{1}{c|}{\begin{tabular}[c]{@{}c@{}}CASIA-\\ MFSD\end{tabular}} & \begin{tabular}[c]{@{}c@{}}\textbf{SuHiFi-}\\ \textbf{Mask(Ours)}\end{tabular} \\ \hline
\multicolumn{2}{|c|}{ResNet18}                        & \multicolumn{1}{c|}{36.3}                                                     & \textbf{44.5}                                                   & \multicolumn{1}{c|}{\textbf{50.9}}                                                  & 42.1                                                   \\ \hline
\multicolumn{2}{|c|}{ViT}                             & \multicolumn{1}{c|}{34.9}                                                    & \textbf{42.8}                                                   & \multicolumn{1}{c|}{44.8}                                                  & \textbf{45.9}                                                   \\ \hline
\multicolumn{2}{|c|}{CDCN}                            & \multicolumn{1}{c|}{15.6}                                                     & \textbf{45.9}                                                   & \multicolumn{1}{c|}{32.6}                                                  & \textbf{41.4}                                                   \\ \hline
\multicolumn{2}{|c|}{AUX.(Depth)}                     & \multicolumn{1}{c|}{27.6}                                                     & \textbf{43.8}                                                   & \multicolumn{1}{c|}{28.4}                                                  & \textbf{39.6}                                                   \\ \hline
\end{tabular}
}
\label{table:crossdata}
\end{table}

\begin{table}[ht]
\centering
\caption{Cross-testing results on the MARsV2 degraded with different size of the gaussian kernel when trained on the proposed SuHiFiMask.}
\setlength{\tabcolsep}{1mm}
\scalebox{0.8}{
\begin{tabular}{|c|c|c|c|c|c|c|c|}
\hline
\multicolumn{1}{|c|}{\multirow{3}{*}{Method}} & Train  & \multicolumn{6}{c|}{\textbf{SuHiFiMask (ours)}}                                                                                                                                 \\ \cline{2-8} 
\multicolumn{1}{|c|}{}                        & Test   & \multicolumn{2}{c|}{MARsV2}                                  & \multicolumn{2}{c|}{MARsV2-3×3}                            & \multicolumn{2}{c|}{MARsV2-5×5}       \\ \cline{2-8} 
\multicolumn{1}{|c|}{}                        & Metric & \multicolumn{1}{c|}{HTER(\%)${\downarrow}$} & \multicolumn{1}{c|}{AUC(\%)${\uparrow}$} & \multicolumn{1}{c|}{HTER (\%)${\downarrow}$} & \multicolumn{1}{c|}{AUC (\%)${\uparrow}$} & \multicolumn{1}{c|}{HTER(\%)${\downarrow}$} & AUC (\%)${\uparrow}$ \\ \hline
\multicolumn{2}{|c|}{ResNet18}                         & \multicolumn{1}{c|}{27.2}     & \multicolumn{1}{c|}{79.6}    & \multicolumn{1}{c|}{29.5}     & \multicolumn{1}{c|}{79.2}    & \multicolumn{1}{c|}{32.5}     & \textbf{75.5}    \\ \hline
\multicolumn{2}{|c|}{CDCN}                             & \multicolumn{1}{c|}{37.6}     & \multicolumn{1}{c|}{66.8}    & \multicolumn{1}{c|}{41.6}     & \multicolumn{1}{c|}{61.7}    & \multicolumn{1}{c|}{51.2}     & 52.7    \\ \hline
\multicolumn{2}{|c|}{AUX.(Depth)}                      & \multicolumn{1}{c|}{26.8}     & \multicolumn{1}{c|}{79.4}    & \multicolumn{1}{c|}{41.1}     & \multicolumn{1}{c|}{63.7}    & \multicolumn{1}{c|}{48.7}     & 54.5    \\ \hline
\multicolumn{2}{|c|}{\textbf{CQIL (ours)}}                       & \multicolumn{1}{c|}{\textbf{21.8}}     & \multicolumn{1}{c|}{\textbf{87.5}}    & \multicolumn{1}{c|}{\textbf{26.2}}     & \multicolumn{1}{c|}{\textbf{81.4}}    & \multicolumn{1}{c|}{\textbf{30.9}}     & 74.0    \\ \hline
\end{tabular}
}
\label{table:crossmethod}
\end{table}

\textbf{Experiments on Protocol 3-Quality.}
Protocol 3 evaluates the stability of the algorithm to image quality degradation. Since the training, development, and testing sets of this protocol differ only in quality, the algorithm is needed to learn a general feature extraction method on data with different quality distributions. As shown in Tab.~\ref{table:Protocol1}, our algorithm ranks first on protocol 3 (APCER, BPCER, and ACER are 19.14\%, 12.82\%, 15.98\%, respectively), which proves that our algorithm is effective in extracting discriminative features independent of quality.

\subsection{Inter-Testing.} To evaluate the difficulty of surveillance-based FAS tasks and the effectiveness of CQIL working on low-quality datasets, we design a number of cross-testing experiments.
%crossmethod

\textbf{Cross-dataset.} To further evaluate the difficulty of the long-distance PAD task based on surveillance scenes, we design two cross-dataset experiments. (1) We train the model on the CASIA-MFSD dataset and perform the cross-test evaluation on the proposed SuHiFiMask and ReplayAttack datasets. (2) We train the model on the ReplayAttack dataset and evaluate it on the SuHiFiMask and CASIA-MFSD datasets for cross-testing. As shown in Tab.~\ref{table:crossdata}, the performance of the model tested on the proposed SuHiFiMask is significantly degraded relative to the performance testing on ReplayAttack or CASIA-MFSD. For example, the HTER (\%) of the CDCN trained on CASIA-MFSD was increased by $30.3\%$ for the test on the proposed dataset compared to the test on ReplayAttack. This shows the performance of existing algorithms degrades significantly when they encounter negative factors such as low resolution, motion blur, and occlusion. In particular, CQIL is a PAD method based on low-quality data and requires low-resolution images as input. So the generality of the method will be evaluated in the next subsection.

\begin{figure}[ht]
	\centering
	\includegraphics[width=1.0\linewidth]{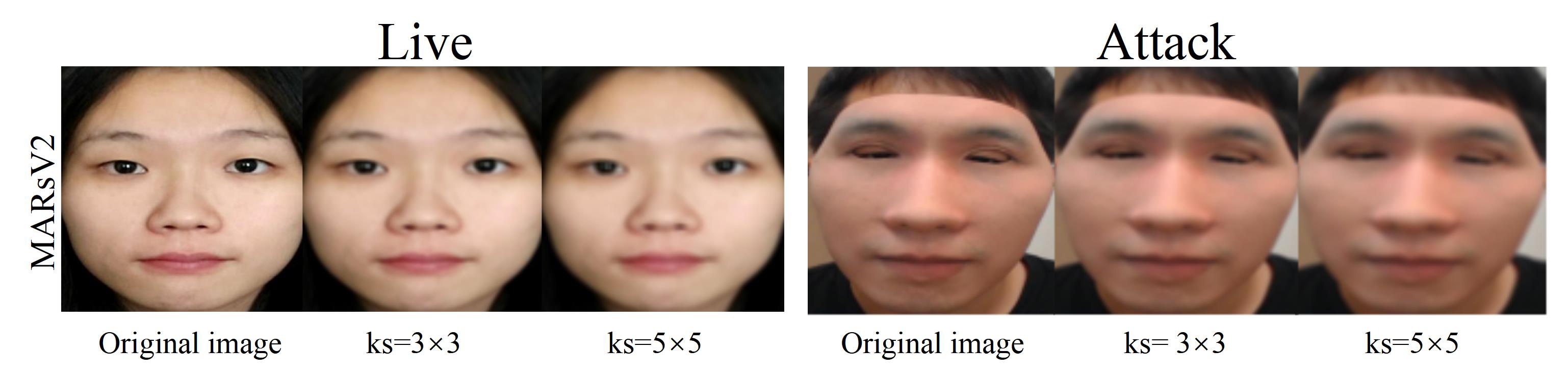}
	\vspace{-0.4cm}
	\caption{MARsV2 with different sizes of the gaussian kernel processing.}
	\label{fig:mars}
\end{figure}

\textbf{Cross-quality.} To demonstrate the generality of our method to low-quality datasets, we design a series of experiments across the quality. We train the different methods on the proposed SuHiFiMask and test them on MARsV2 after the degradation of gaussian kernels of different sizes. Specifically, since no existing work has provided available low-quality PAD datasets, we simulate low-quality datasets with different degrees of degradation by means of a gaussian kernel to verify the generality of different methods on low-quality datasets. Fig.~\ref{fig:mars} shows several samples of MARsV2 after treatment with gaussian kernels of different sizes. As shown in Tab.~\ref{table:crossmethod}, our CQIL achieves good performance on the MARsV2 dataset at all degradation degrees. This demonstrates that our method can encode quality-independent discriminative features. However, there is a domain gap between the manually degraded low-quality dataset and the dataset based on the surveillance scenes. This results in CQIL not being able to take full advantage of encoders trained on low-quality data in real surveillance scenes.

%marsv2退化

\subsection{Visualization Analysis}
\begin{figure}[ht]
	\centering
	\includegraphics[width=0.9\linewidth, height=0.45\textwidth]{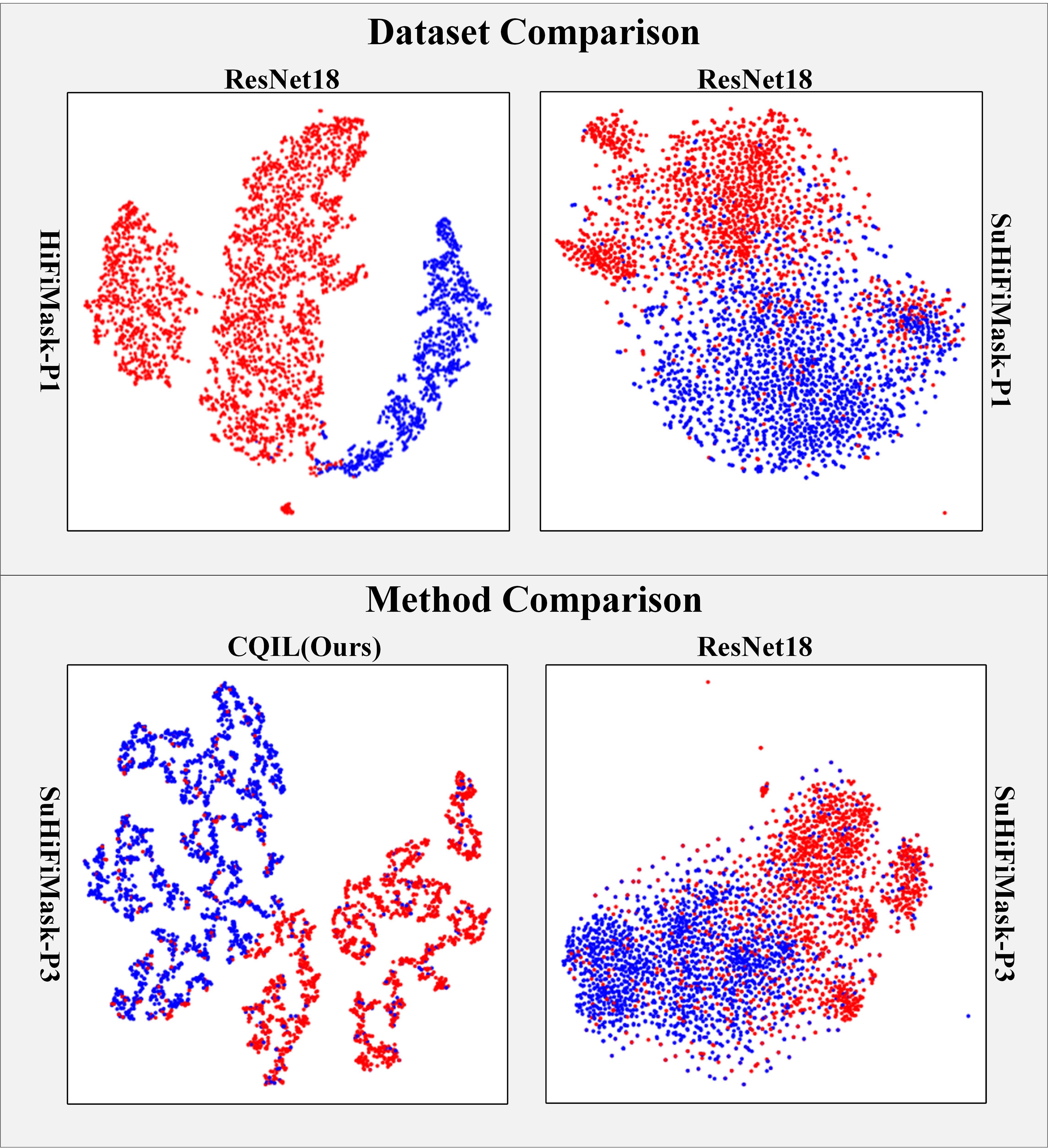}
	\vspace{-0.4cm}
	\caption{Feature distribution comparison on HiFiMask and SuHiFiMask using t-SNE~\cite{van2008visualizing}. The points with different colors denote features from different classes (blue: real faces; red: attack samples).}
	\label{fig:visionfeat}
\end{figure}

In this section, we further visualize the difficulties that low-quality data poses to FAS work and the performance of CQIL in surveillance scenes. First, we compare the features learned by ResNet18 on protocol 1 of HiFiMask, a dataset for the constrained environment, and on protocol 1 of SuHiFiMask, a proposed surveillance scene-based dataset. As shown in Fig~\ref{fig:visionfeat}, the performance of the algorithm degrades significantly on SuHiFiMask, which indicates that the low-quality data in the surveillance scenes add difficulties to the FAS work. Next, we compare the features learned by CQIL and ResNet18 on protocol 3 of the proposed SuHiFiMask. Compared with ResNet18, the proposed CQIL is able to better distinguish between real faces and attacks, which demonstrates the better discriminative representation capacity of the proposed CQIL in surveillance scenes.

%vision

\section{Conclusion}
In this paper, we release the first large-scale FAS dataset based on surveillance scenes, SuHiFiMask, with three challenging protocols. We hope that this will fill the gap in FAS research in long-distance surveillance scenes. In addition, we propose a Contrastive Quality-Invariance Learning (CQIL) network to recover image information using super-resolution and enhance the robustness of the algorithm to quality variations by fitting the quality variance distribution. Finally, we conduct comprehensive experiments on SuHiFiMask and three other datasets to verify the importance of the datasets for the FAS task and the effectiveness of the proposed method.

\ifCLASSOPTIONcaptionsoff
  \newpage
\fi
\bibliographystyle{IEEEtran}
\bibliography{IEEEabrv,reference}

\newpage

\titleformat{\section}[display]{\bfseries\centering}{Appendix}{1em}{}
\section*{Appendix}

\subsection{Sample of faces}
\begin{figure*}[htp]
	\centering
	\includegraphics[width=1.0\linewidth, height=1.2\textwidth]{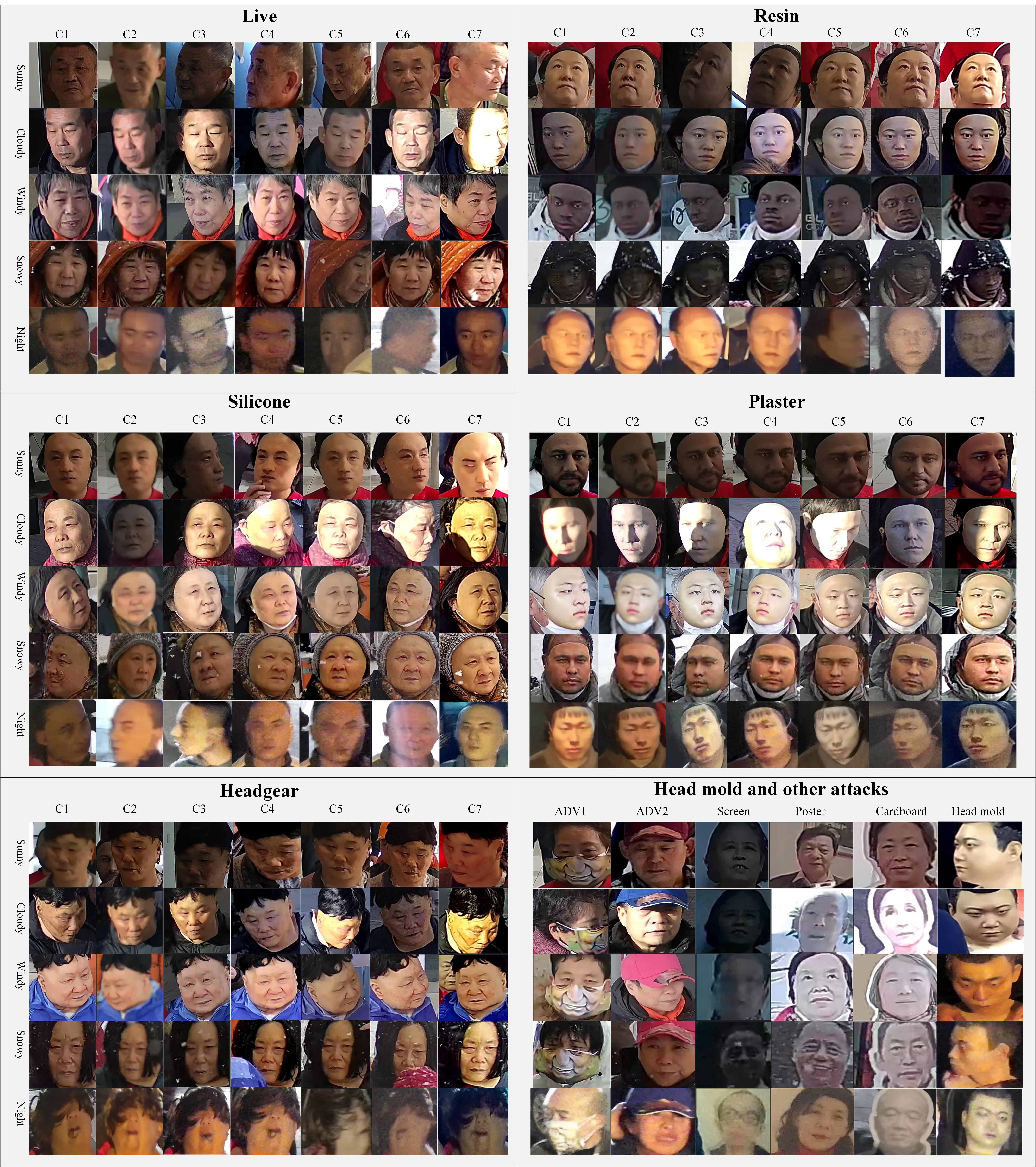}
	\vspace{-0.4cm}
	\caption{We show some samples taken on snowy, cloudy, windy, sunny and night time days. Among them, C1 to C7 represents the surveillance cameras with DS-2CD3T87WD-L, DS-2CD3T86FWDV2-I3S, TL-IPC586HP, TL-IPC586FP, DH-IPC-HFW4843M, DH-P80A1-SA, and ZD5920-Gi4N, respectively.}
	\label{fig:face}
\end{figure*}

As shown in the Fig.~\ref{fig:face}, we have listed some samples of pre-processed face images. The figure contains six sections, which are listed as close-up samples of real people, resin masks, silicone masks, plaster masks, headgear, head molds and other forms of attacks. Each column in the first five sections of the figure represents a mainstream surveillance camera, where C1 to C7 represents DS-2CD3T87WD-L, DS-2CD3T86FWDV2-I3S, TL-IPC586HP, TL-IPC586FP, DH-IPC-HFW4843M, DH-P80A1-SA, and ZD5920-Gi4N cameras. Each row in the first five sections of the figure represents a weather or shooting time, with samples taken on sunny days, cloudy days, windy days, snowy days, and nights, respectively. The sixth section of the figure lists head molds and other forms of attack, from left to right, in each column are adversarial masks, adversarial hats, replay attacks in electronic screens, posters, cardboards, and head molds.

\subsection{Sample of scenes}
\begin{figure*}[ht]
	\centering
	\includegraphics[width=1.0\linewidth, height=1\textwidth]{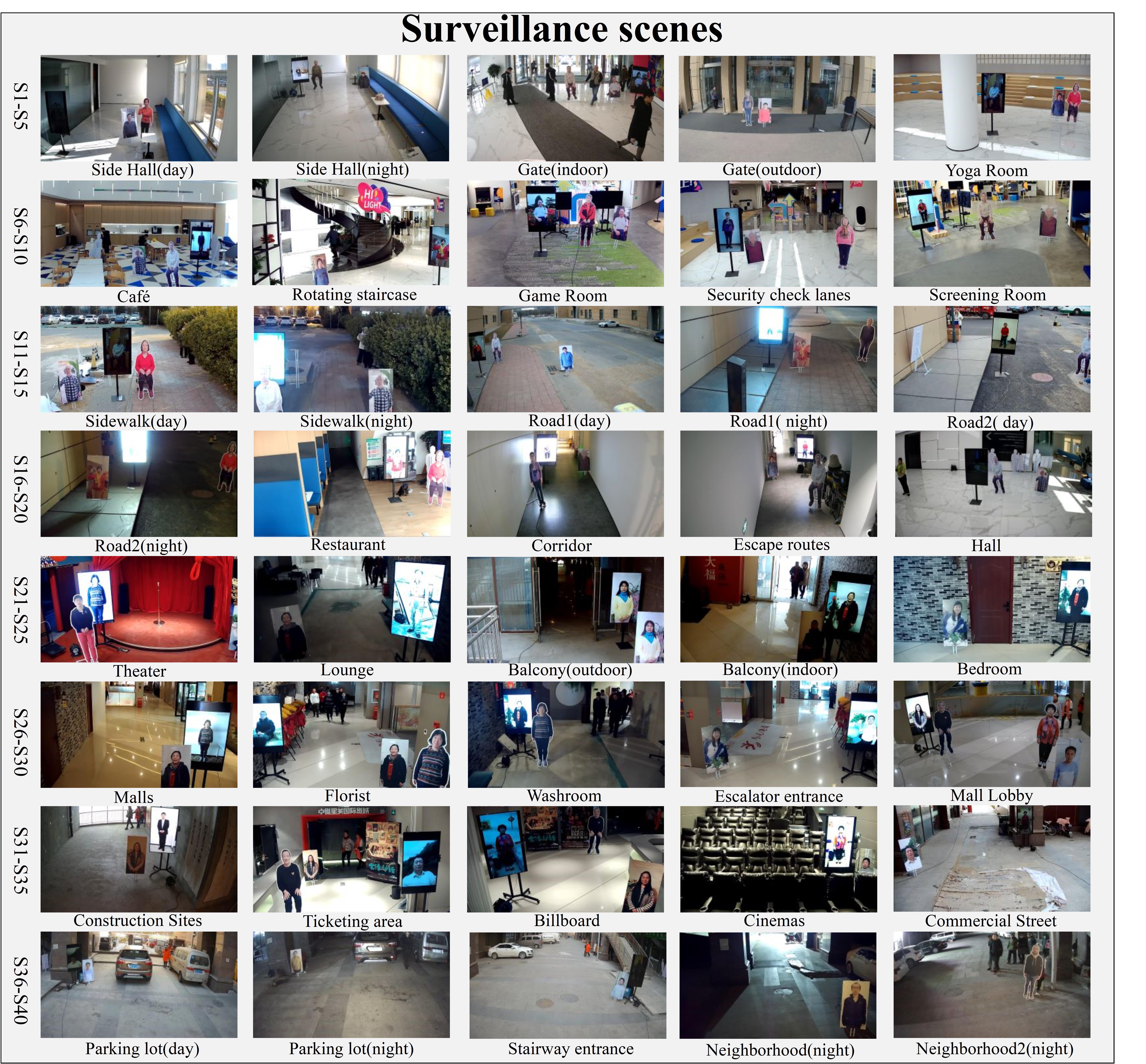}
	\vspace{-0.4cm}
	\caption{We show the 40 surveillance scenes included in SuHiFiMask, S1-S40 on the left side of the figure are the numbers of each scene.}
	\label{fig:scene}
\end{figure*}

As shown in the Fig.~\ref{fig:scene}, we have listed all 40 scenes included in the SuHiFiMask, which include daily life scenes (\eg, cafes, cinemas, and theaters) and security check scenes (\eg, security check lanes and parking lots) for deploying face recognition systems. On the left side of the figure is the number of each scene in the row, which is the basis for naming the videos in the dataset. In addition, we need to increase the relevance of the data content and surveillance scenes by asking the subjects do scene-related behaviors in the scenes, such as asking the subjects sit around a coffee table and drink coffee in the coffee shop scenes. It is worth mentioning that some scenes in real life are vulnerable to attack in both day and night, so we identify the day and night of this scene as two different scenes, such as parking lot (day) and parking lot (night).

\end{document}